%% file: egpaper_for_review.tex
\documentclass[10pt,twocolumn,letterpaper]{article}

\usepackage{cvpr}
\usepackage{times}
\usepackage{epsfig}
\usepackage{graphicx}
\usepackage{amsmath}
\usepackage{amssymb}
\usepackage{subfig}
\usepackage{comment}
\usepackage{caption}
\usepackage{wasysym}
\usepackage{amsthm}
\usepackage{color}
\usepackage{float}
\usepackage{multirow}
\usepackage{graphics}
\usepackage[english]{babel}
\usepackage{titling}

\usepackage[pagebackref=true,breaklinks=true,letterpaper=true,colorlinks,bookmarks=false]{hyperref}
\usepackage[titlenumbered,ruled]{algorithm2e}

\cvprfinalcopy 

\def\cvprPaperID{1409} 

\ifcvprfinal\fi

\newtheorem{theorem}{Theorem}[]
\newtheorem{corollary}{Corollary}[]
\newtheorem{lemma}[theorem]{Lemma}

\newtheorem{definition}{Definition}[]
\newcommand{\Real}{\mathbb R}

\PassOptionsToPackage{normalem}{ulem}
\usepackage{ulem}
\usepackage{babel}

\newcommand{\vv}{\mathbf{v}}
\newcommand{\vt}{\mathbf{t}}

\newcommand{\rank}{\text {rank}}

\definecolor{mypink1}{rgb}{0.33, 0.76, 0.66}
\newcommand{\yk}[1]{\textcolor{mypink1}{[Yoni: #1]}}
\newcommand{\rb}[1]{\textcolor{magenta}{[Ronen: #1]}}

\newcommand{\ag}[1]{\textcolor{red}{[Amnon: #1]}}

\date{}
\begin{document}

\title{GPSfM: Global Projective SFM Using Algebraic Constraints\\ on Multi-View Fundamental Matrices}
\author{Yoni Kasten* \hspace{1cm} Amnon Geifman* \hspace{1cm} Meirav Galun  \hspace{1cm}Ronen Basri \\
Weizmann Institute of Science\\
{\tt\small  \{yoni.kasten,amnon.geifman,meirav.galun,ronen.basri\}@weizmann.ac.il}
}

\maketitle

\begin{abstract}
  \input{abstract}

\end{abstract}
\makeatletter
\def\blfootnote{\xdef\@thefnmark{}\@footnotetext}
\makeatother
\blfootnote{*Equal contributors}
\section{Introduction}
\input{introduction}



\section{Algebraic constraints of $n$-view fundamental matrices}
\label{sec:theory}
\input{theoretical_part}

\section{Method}
\input{method}

\input{method2}

\section{Experiments}
\input{experiments}

{\small
\bibliographystyle{ieee}
\bibliography{egbib}
}

\include{SM}



\end{document}

%% file: abstract.tex
This paper addresses the problem of recovering projective camera matrices from collections of fundamental matrices in multiview settings. We make two main contributions. First, given ${n \choose 2}$ fundamental matrices computed for $n$ images, we provide a complete algebraic characterization in the form of conditions that are both necessary and sufficient to enabling the recovery of camera matrices. These conditions are based on arranging the fundamental matrices as blocks in a single matrix, called the $n$-view fundamental matrix, and characterizing this matrix in terms of the signs of its eigenvalues and rank structures. Secondly, we propose a concrete algorithm for projective structure-from-motion that utilizes this characterization. Given a complete or  partial collection of measured fundamental matrices,  our method seeks camera matrices that minimize a global algebraic error for the measured fundamental matrices. In contrast to existing methods, our optimization, without any initialization,  produces a consistent set of fundamental matrices that corresponds to a unique set of cameras (up to a choice of projective frame).  Our experiments indicate that our method achieves state of the art performance in both accuracy and running time.

%% file: introduction.tex
This paper considers the problem of recovering projective camera matrices from collections of fundamental matrices. Many multiview structure from motion (SFM) pipelines begin, given $n$ images, $I_1,...,I_n$, by robustly estimating fundamental matrices between image pairs from collections of point matches, e.g., using RANSAC. However, in typical settings, only a subset of the ${n \choose 2}$ pairwise fundamental matrices can be estimated, and the estimated matrices may be subject at times to significant errors. Moreover, fundamental matrices are defined through a homogeneous equation and can  thus assume any scale factor, but a consistent setting of these scales is important in multiview settings \cite{rudi2010linear, sengupta2017new} (in analogy to resolving\ the distance between cameras in a calibrated setting).  Consequently, accurate recovery of camera matrices is interesting both from theoretical and practical standpoints. Our paper makes contributions to both of these aspects.

An important theoretical question is, given a collection of ${n \choose 2}$ fundamental matrices, whether these fundamental matrices are \textit{consistent}, in the sense that there exist $n$ camera matrices that produce these fundamental matrices. Below we address this question by providing a set of algebraic constraints that form both necessary and sufficient conditions for the consistency of fundamental matrices. Our formulation, which extends  the partial list of necessary conditions introduced in \cite{sengupta2017new}, considers the symmetric matrix $F$ of size $3n \times 3n$, formed by stacking all $n \choose 2$ fundamental matrices. It provides a complete characterization of $F$ in terms of the signs of its eigenvalues and rank patterns. 

An advantage of our algebraic characterization is that it can readily be used to construct optimization algorithms to recover camera matrices. In the second part of this paper we introduce an efficient algorithm to recover projective camera matrices directly from measured fundamental matrices. Our algorithm, which utilizesã the consistency constraints presented in this paper, uses global optimization for camera recovery, overcoming noise and missing measurements. It further
avoids one of the main difficulties in a previous approach \cite{sengupta2017new} -- the need to accurately recover a scale factor for each of the estimated fundamental matrix. This allows us to obtain state-of-the-art results without any initialization. We  demonstrate the utility of our method by applying it to uncalibrated image collections of various sizes. Our experiments indicate that our method outperforms previous methods in both accuracy and runtime.

\subsection{Previous work}

The recovery of projective camera matrices  was addressed in several lines of work.

\noindent \textbf{\uline{Incremental algorithms}} \cite{klopschitz2010robust,magerand2017practical,pollefeys2004visual} process the images sequentially, interleaving camera and depth recovery for every new image. Such methods can be sensitive to the order of processing and may suffer from drift, due to accumulation of errors. The use of bundle adjustment for every new image reduces such drift, but is computationally demanding.

\noindent \textbf{\uline{Factorization-based methods}} \cite{christy1996euclidean,dai2013projective,kennedy2016online,magerand2017practical,oliensis2007iterative,sturm1996factorization,ueshiba1998factorization} factor a measurement matrix that includes all point matches across views into an (unknown) product of camera matrices, depth values, and 3D point locations. These methods typically yield very large optimization problems and are often approached by splitting the problem into smaller subproblems.

\noindent \textbf{\uline{Global methods}}. A number of ``global methods" were proposed recently demonstrating both accurate and efficient recovery of camera matrices from pairwise measurements (essential or fundamental matrices), mostly in a calibrated setting \cite{jiang2013global,wilson2014robust,ozyesil2015robust,goldstein2016shapefit,arie2012global}. Sweeney el al. \cite{sweeney2015optimizing} attempts to improve the consistency of fundamental matrices by minimizing the discrepancy of reprojected points in three views (through an ``epipolar point transfer"). They, however, cannot achieve projective recovery since their method does not guarantee the consistency of the improved fundamental matrices. Sengupta et al. \cite{sengupta2017new} attempt to enforce rank constraints on the measured fundamental matrices. Complicated by the need to simultaneously recover suitable scale factors, their method is sensitive to errors and requires highly accurate initialization, which was achieved by applying the state-of-the-art, \textit{calibrated} LUD algorithm \cite{ozyesil2015robust}, defeating the purpose of projective camera recovery without calibration. 

\noindent \textbf{\uline{Solvability of viewing graphs}}. A number of papers seek to design algorithms that can identify ``solvable viewing graphs" \cite{levi2003viewing,ozyesil2015robust,rudi2010linear,sweeney2015optimizing}. A \textit{viewing graph} captures the pattern of missing fundamental matrices. Let $G=(V,E)$ be a graph such that a node $v_i \in V$ represents image $I_i$ and an edge $e_{ij} \in E$ exists if the fundamental matrix $F_{ij}$ relating $I_i$ and $I_j$ is available. $G$ is called \textit{solvable} if the corresponding $n$ camera matrices can be determined uniquely (up to a $4 \times 4$ projective transformation) despite the missing  fundamental matrices. Identifying solvable graphs is equivalent to asking, given a partial set of fundamental matrices, if $F$ can be completed uniquely to satisfy our algebraic constraints.

Finally, both our paper and  \cite{sengupta2017new} explore algebraic properties of multiview fundamental matrices (MVFs) and propose optimization schemes for utilizing these properties toward camera pose recovery. However, these two papers differ in several significant respects. First, \cite{sengupta2017new} provides a set of  necessary algebraic constraints, for the consistency of MVF $F$, while we provide a complete set of necessary and sufficient algebraic conditions.  Moreover, our conditions are specified directly in terms of the MVFs, in  contrast to \cite{sengupta2017new} which rely on the construction of an auxiliary, unknown matrix. Our direct formulation further leads to a significantly simpler optimization algorithm,  including a new algorithm for recovering the projective camera matrices, from a consistent MVF $F$, which is lacking in \cite{sengupta2017new}

%% file: theoretical_part.tex
Let $I_1, \ldots, I_n$ denote a collection of $n$ images of a static scene captured respectively by projective cameras $P_1,...,P_n$. Each camera $P_i$ is represented by a $3 \times 4$ matrix $P_i= K_i R_i^T[I,-{\bf t}_i]$, where $K_i$ is a $3 \times 3$ calibration matrix, ${\bf t}_i \in \Real^3$ and $R_i \in SO(3)$ respectively denote the location and orientation of $P_i$ in some global coordinate system, and $I$ denotes the $3 \times 3$ identity matrix. Below we further denote $V_i = K_i^{-T}R_i^T$, so $P_i= V_i^{-T}[I,-{\bf t}_i]$. Consequently, let $\bold{X}=(X,Y,Z)^T$ be a scene point in the global coordinate system. Its projection onto $I_i$ is given by ${\bf x}_i = \bold{X}_{i} / Z_i$, where $\bold{X}_{i}=(X_i, Y_i, Z_i)^T = K_{i}R_{i}^{T}(\bold{X}-{\bf t}_{i})$.

We next denote the fundamental matrix between images $I_i$ and $I_j$ by $F_{ij}$. In  \cite{arie2012global,sengupta2017new}  it was shown that $F_{ij}$ can be written as 
\begin{equation}  \label{eq:fundamental}
F_{ij} = K_i^{-T}R_i^T (T_i - T_j) R_j K_j^{-1} = V_i(T_i-T_j)V_j^T,
\end{equation}
where $T_i=[{\bf t}_i]_{\times}$. It can be readily verified that this definition of $F_{ij}$ is consistent with the standard properties of fundamental matrices~\cite{hartley2003multiple}, including (a) $P_{i}^{T}F_{ij}P_{j}$ is skew symmetric, and (b) ${\bf e}^T_{ik} F_{ij} {\bf e}_{jk} =0$, where ${\bf e}_{ik}$ denotes the projection of the center of camera $k$ onto camera $i$, i.e., ${\bf e}_{ik} = P_i {\bf t}_k$. 

Note, however, that \eqref{eq:fundamental} attributes a scale to each $F_{ij}$ that relates it to some global coordinate system. In practice, given two images, a fundamental matrix is determined only up to a multiplicative factor. We will denote an estimated fundamental matrix by $\hat F_{ij}$ and assume, in case it is estimated accurately, that $\hat F_{ij}=\lambda_{ij}F_{ij}$ with unknown $\lambda_{ij} \ne 0$. 

We next construct a matrix from all $n \choose 2$ fundamental matrices. 

\begin{definition}  \label{def:mv_fundamental} 
A matrix $F \in \mathbb{S}^{3n}$, whose $3 \times 3$ blocks are denoted by $F_{ij}$, is called an $n$-view fundamental matrix if $rank(F_{ij})=2$ for all $i \ne j$ and $F_{ii}=0$.
\end{definition}
\noindent We use $\mathbb{S}^{3n}$ to denote the space of all $3n \times 3n$ symmetric matrices. The symmetry of \(F\) implies that $F_{ij}=F_{ji}^T$.
\begin{definition}  \label{def:mv_fundamental} 
An $n$-view fundamental matrix $F$ is called consistent if there exist camera matrices $P_1,...,P_n$ of the form $P_{i}=V_{i}^{-T}[I,\mathbf{t}_{i}]$ such that $F_{ij}=V_{i}([\mathbf{t}_{i}]_{\times}-[\mathbf{t}_{j}]_{\times})V_{j}^{T}$.
\end{definition}
\noindent A consistent $F$, therefore, takes the form
\begin{align*}
F & =\left(\begin{array}{cccc}
0 & F_{1 2} & ... & F_{1 n}\\
F_{2 1} & 0 & ...& F_{2n}\\
\vdots & & & \vdots\\
F_{n1} & F_{n2} & ... & 0
\end{array}\right)
\end{align*}
and each $F_{ij}$ is scaled properly in accordance with the global coordinate system. We further refer to the  $3 \times 3n$ $i^{th}$ block row of $F$ by $F_i$.

Our main theoretical result is summarized below in Theorem~\ref{thm:consistent_direct}, which specifies a set of necessary and sufficient algebraic conditions for the consistency of $F$ in terms of its eigenvalue sign and rank patterns. These, in turn, will be used in later sections to construct a new optimization algorithm for global recovery of  projective camera matrices from noisy fundamental matrices.

\begin{theorem}\label{thm:consistent_direct} 

An n-view fundamental matrix $F$ is consistent with a set of $n$ cameras whose centers are not all collinear if, and only if, the following conditions hold: 
\begin{enumerate}
\item $Rank(F)=6$ and $F$ has exactly 3 positive and 3 negative eigenvalues.
\item $Rank(F_i) = 3$ for all $i =1, ..., n$.
\end{enumerate}
\end{theorem}

\noindent Below we provide a proof sketch. The full proof is deferred to the supplementary material. To prove the theorem we first state that for a symmetric rank 6 matrix $F$ the following three conditions are equivalent:
\begin{enumerate}
\item[(i)] $F$ has exactly 3 positive and 3 negative eigenvalues.
\item[(ii)] $F=XX^T-YY^T$ with $X,Y \in \Real^{3n \times 3}$ and $rank(X)=rank(Y)=3.$
\item[(iii)] $F=UV^T+VU^T$ with $U,V \in \Real^{3n \times 3}$ and $rank(U)=rank(V)=3$.
\end{enumerate} 
In particular, using the eigen-decomposition of $F$, let $F\boldsymbol x_i=\alpha_i \boldsymbol x_i$ and $F\boldsymbol y_i=-\beta \boldsymbol y_i$, $\alpha_i,\beta_i > 0$, the columns of $X$ and $Y$ respectively may include $\sqrt{\alpha_i}\boldsymbol x_i$ and $\sqrt{\beta_i}\boldsymbol y_i$, and $U,V$ are related to $X,Y$ through
\begin{equation}  \label{eq:xy2uv}
U=(X-Y)/\sqrt{2}, ~~~ V=(X+Y)/\sqrt{2}.
\end{equation}

Next, to show the necessary condition, let $F$ be a consistent, $n$-view fundamental matrix, then clearly \eqref{eq:fundamental} can be written in matrix form as $F=UV^T+VU^T$, where $U,V \in \Real^{3n \times 3}$ whose $3 \times 3$ blocks respectively are $U_i=V_iT_i$ and $V_i$, implying condition 1. Condition 2 holds because not all cameras are collinear, since if conversely  $rank(F_i) < 3$ for some $i$ then there exists a 3-vector ${\bf e} \neq 0$ such that $F_i^T {\bf e}=0$, and therefore $\forall j~F_{ji} {\bf e}=0$, implying, in contradiction, that the   camera centers are all collinear.

To establish the sufficient condition, let $F$ be an $n$-view fundamental matrix that satisfies conditions 1 and 2. Condition 1 (along with (iii)) implies that $F_{ij}=U_{i}V_{j}^{T}+V_{i}U_{j}^{T}$. Next, $F_{ii}=0$ implies that  $U_{i}V_{i}^{T}$ is skew symmetric, and so $\forall i$ either $rank(U_i)=2$ or $rank(V_i)=2$. Next, as we show in the supplementary material, $rank(F_i)=3$  implies WLOG that $\forall i, rank(V_{i})=3$ and $rank(U_{i})=2$. This and the skew-symmetry of $U_iV_i^T$ imply that $V_i^{-1}U_i$ is skew-symmetric. Denote this matrix by $T_i=[{\bf t}_i]_{\times}$, we obtain $F_{ij}=V_{i}(T_{i}-T_{j})V_{j}^{T}$, establishing that $F$ is consistent. Finally, $\{{\bf t}_i\}_{i=1}^n$ are not all collinear, since, otherwise $\exists i$ and $\exists {\bf e} \neq 0$ such that $\forall j~F_{ji} {\bf e}=0$, implying that $F_i^T {\bf e}=0$, contradicting the full rank of $F_i$.

Theorem \ref{thm:consistent_direct} also provides a practical tool for projective reconstruction. Given a set of (possibly noisy) pairwise fundamental matrices we can use constrained, low-rank optimization to recover a matrix that  satisfies conditions 1-2. Then, we can use the obtained $n$-view fundamental matrix to recover the underlying camera matrices. This is summarized in the following corollary.

\begin{corollary}
\label{corollary:projective_reconstruction} Projective reconstruction: Let $F \in \mathbb{S}^{3n}$ be a consistent $n$-view fundamental matrix, then it is possible to explicitly determine camera matrices $P_1,...,P_{n}$ that are consistent with $F$. 
\end{corollary}

\begin{proof}
The claim is justified by the following construction, which relies on the proof of Theorem~\ref{thm:consistent_direct}. 
\begin{enumerate}
\item Since $F$ satisfies condition 1 we can use its eigen-decomposition to express it as $F=XX^{T}-YY^{T}$, and then construct $U$ and $V$ using \eqref{eq:xy2uv}.
\item Now, WLOG, $rank(V_{i})=3$ and $rank(U_{i})=2$ for all $i=1,...,n$ (or else $U$ and $V$ should be interchanged).
\item We next define $T_i=V_i^{-1}U_{i}$. $T_{i}$ is skew symmetric, and we denote $T_i=[{\bf t}_i]_\times$.

\item  By construction $F_{ij}= V_i(T_i - T_j)V_j^T$, implying that $P_{i}=[V_{i}^{-T}|-V_{i}^{-T}{\bf t}_{i}]$
form a consistent set of camera matrices.
 
 
%
%
\end{enumerate}
This construction is unique up to a $4 \times 4$ projective transformation.
\end{proof}

Let $F$ be a consistent $n$-view fundamental matrix. Clearly, if we scale differently any of its blocks, $\lambda_{ij}F_{ij}$ with $\lambda_{ij} \ne 0$, then in general $F$ ceases to be consistent. In particular, it maintains condition 2 of Theorem~\ref{thm:consistent_direct}, but its rank is no longer 6. Note, however, that we can scale each block-row (and by symmetry column) of $F$ differently and maintain both the conditions of the theorem; i.e., let $S=diag(s _1I,..,s _nI)$ with $s_i \ne 0$, then $SFS$ is consistent if and only if $F$ is consistent. (Such scaling is equivalent to scaling the projective camera matrices.) This, in fact, implies that we need to determine only ${n \choose 2} - n$ of the scale factors and set the rest of the $n$ scales arbitrarily.

Up to this point  we considered $n$-view matrices that include all $n \choose 2$ pairwise fundamental matrices. In real applications, however, often only a subset of the pairwise matrices can be computed. Additionally, the estimated fundamental matrices are improperly scaled and may suffer from large inaccuracies. In these cases we may want to reconstruct the cameras from partial subsets of fundamental matrices. Indeed, our algorithm, presented later in Sec.~\ref{section::method}, recovers a consistent set of camera matrices from triplets of images, allowing us to handle missing fundamental matrices and to remove outliers. The following theorem establishes that, with proper intersection, camera matrices are determined uniquely (up to the usual $4 \times 4$ projective ambiguity) from consistent sub-matrices. We first need the following definition:

\begin{definition}
Let $F \in \Real^{3n \times 3n}$ and let \(F^{1},...,F^k\) be block sub-matrices of $F,$ with $F^i \in \Real^{3m_i \times 3m_i}$, $3 \le m_i \le n$. \(\{F^{1},...,F^k\}\) is called a \textbf{consistent cover} of \(F\) if each \(F^i\) forms a consistent multi-view fundamental matrix and each  diagonal element  of $F$ is contained in at least one of \(F^{1},...,F^k\).
\end{definition}


\begin{theorem} \label{thm:merging}
Let $F \in \Real^{3n \times 3n}$ and  let \(F^{1},...,F^k\) form a consistent cover of $F$. If for all \(2 \le m \le k\), there exists \(l < m\) such that \(F^l,F^m\) overlap in at least one fundamental matrix, then there exists a unique \(n\)-view consistent fundamental matrix $\bar{F}$ (up to \(n\) scale factors) whose blocks \(\bar F_{ij}=\lambda_{ij}F_{ij}\) with some \(\lambda_{ij} \ne 0\) for all $F_{ij}$ that belong to any of \(F^1, ..., F^k\).

\end{theorem}
\begin{proof}
We prove this by induction on \(k\). We begin with \(k=2\). By Corollary \ref{corollary:projective_reconstruction}, $F^1,F^2$ define two sets of camera matrices ${\cal P}^1,{\cal P}^2$ that are consistent with respect to $F^1, F^2$, respectively. Since $F^1$ and $F^2$ share a fundamental matrix $F_{ij}$,  $F_{ij}$ corresponds to a pair of cameras in  ${\cal P}^1$ and a second pair in ${\cal P}^2$ so that the two pairs are equal up to a projective homography (\cite{hartley2003multiple}, p.\ 254). Consequently, ${\cal P}^2$ can be mapped to the projective frame of ${\cal P}^1$ to form a set of $n$ camera matrices \cite{rudi2010linear}, that in turn determine a unique (up to $n$ global scale factors) consistent $n$-view matrix, $\bar{F}$. Now, each fundamental matrix, in either $F^1$ or $F^2,$  corresponds to two cameras from this set of  $n$ cameras and hence has exactly the same entries as in $\bar{F}$ up to scale.

This argument can now be repeated inductively to prove the theorem for all $k >2$.
\end{proof}

%% file: method.tex
\label{section::method}
Given images $I_1,...,I_n$, we assume a standard robust method (e.g., RANSAC) is used to estimate the pairwise fundamental matrices, where we denote by $\Omega = \{\hat{F}_{ij}\}$  the set of estimated fundamental matrices.
In general, only a subset of the ${n \choose 2}$ pairwise fundamental matrices are estimated, due to, e.g., occlusion, large motion, or changes in brightness, and the available estimates are noisy. An additional complication is that to make these fundamental matrices consistent they must each be scaled by an unknown factor to fit the global coordinate frame.  Our aim therefore is to find a consistent $n$-view matrix $F \in \mathbb{S}^{3n}$  that is as similar as  possible to the measured fundamental matrices. Straightforward optimization of this problem is difficult as it yields a nonlinear optimization formulation with rank constraints, as in \cite{sengupta2017new}, which required initialization by a high quality method. 

Below we introduce a novel method that utilizes global optimization and yet circumvents the need to recover the scale factors. Our method works by enforcing the consistency constraints of Theorem~\ref{thm:consistent_direct} on 3-view fundamental matrices and by maintaining their intersections, as is required by Theorem~\ref{thm:merging}, to obtain a global reconstruction. This yields an optimization problem, with simple and efficient formulation, that directly uses the measured data without the need to  incorporate any initialization. We solve this optimization using the alternating direction method of multipliers 
(ADMM) \cite{boyd2011distributed}, where each step in the ADMM has a closed form solution.   
  
We avoid recovering scale factors by enforcing consistency for image triplets. As we explained in Sec.~\ref{sec:theory}, we only need to determine ${n \choose 2} - n$ of the scale factors, while the rest can be set arbitrarily. A consequence of this is that there are no scale factors for $n=3$, as proved below.  

\begin{corollary}\label{corollary::scale_factors3} A consistent 3-view fundamental matrix is invariant to arbitrarily scaling its constituent fundamental matrices.
\end{corollary}
\begin{proof}
Let $F$ be a consistent 3-view fundamental matrix whose blocks are defined as $F_{ij}=V_i(T_i-T_j)V_j^T$, and let $\widetilde{F}$ be a $9 \times 9$ matrix whose blocks are defined to be $\widetilde{F}_{ij}=s_{ij}F_{ij}$ where $s_{ij} \ne 0$ are arbitrary scale factors.
Without loss of generality we can assume that the number of negative scale factors is even (otherwise we can multiply the entire matrix by -1). Therefore, $s_1=(\frac{ s_{12}s_{13}}{s_{23}})^{\frac{1}{2}}$,
$s_2=(\frac{ s_{23}s_{12}}{s_{13}})^{\frac{1}{2}}$, and
$s_3 =(\frac{ s_{13}s_{23}}{s_{12}})^{\frac{1}{2}}$ determine real values such
that $s_1s_{2}=s_{12}$, $s_1s_{3}=s_{13}$, and $s_2s_{3}=s_{23}$.
Let $\widetilde{V}_i=s_i V_i$ for $i=1\dots 3$, we get that

\begin{align}
\begin{split}
\widetilde{F}_{ij} = s_{ij} V_i(T_i-T_j) V_j^T
= s_{i} V_i (T_i-T_j) s_{j} V_j^T \\ 
= \widetilde{V}_i (T_i-T_j) \widetilde{V}_j^{T}
\end{split}
\end{align}

Therefore $\widetilde{F} = SFS$ with $S=diag(s_1I,s_2I,s_3I)$,  and hence it is consistent.
\end{proof}

This corollary implies that for 3-view  fundamental matrices consistency is invariant under  \textit{any} choice of scale factors.
Our optimization formulation relies on this  observation to avoid the need to estimate the scale factors during optimization. In particular, we introduce a global optimization scheme that enforces the consistency of triplets of views, while simultaneously enforcing the compatibility of the different triplets. In the rest of this section we first formulate our optimization problem and discuss how to solve it with ADMM. Then we discuss how to select minimal subsets of triplets to speed up the optimization and finally show how the results of our optimization can be used to reconstruct the $n$ cameras.

\subsection{Optimization}
Our input set of estimated fundamental matrices $\{\hat{F}_{ij}\}$ determines a {\it viewing graph} $G=(V,E)$ with nodes $v_1, ..., v_n$,  corresponding to the $n$ cameras, and \(e_{ij} \in E\) if \(\hat F_{ij}\) belongs to the collection of the estimated fundamental matrices. Let $\tau$ denote a collection of $m$ 3-cliques in  $G$, $m \le {n \choose 3}$. The collection $\tau$ may include all the 3-cliques in $G$, or a subset, as we explain in Sec.~\ref{section::graph_construction}. We index the elements of $\tau$ by $k=1,...,m$, where $\tau(k)$ denotes the  \(k^{th}\) triplet.  The selection of $\tau$ induces a partial selection of estimated fundamental matrices, $\Omega$, that participate in the optimization process.  In our construction, if ${\hat F}_{ij} \in \Omega$ then ${\hat F}_{ij}^T = {\hat F}_{ji} \in \Omega$. 


We define the measurement matrix $\hat{F}\in \mathbb{S}^{3n}$ to include all ${\hat F}_{ij} \in \Omega$ in their corresponding $3 \times 3$ block while setting the rest of the blocks to $0_{3 \times 3}$. In the optimization process, we look for a matrix $F \in \mathbb{S}^{3n}$ that is as close as possible to \(\hat F\) under the constraint that  its    $ 9 \times 9$ blocks, induced by $\{\tau(k)\}_{k=1}^m$ and  denoted as $\{F_{\tau(k)}\}_{k=1}^m$, are consistent. In general, such an $F$ is inconsistent (since its scale factors are incompatible across triplets) and incomplete, but, based on Theorem \ref{thm:merging}, it uniquely determines the corresponding projective cameras. 

We next introduce our constrained optimization problem
\begin{align}  \label{eq:opt}
& \min_{F}
& & \sum_{k=1}^m ||F_{\tau(k)}-\hat F_{\tau(k)}||^2_F \\ \nonumber
& \text{s.t.}
& & F=F^T \\ \nonumber
& & &  F_{ii}=0_{3 \times 3} &i=1,\ldots,n \\ \nonumber
& & &  rank(F_{\tau(k)})=6 &k=1,\ldots,m. \nonumber
\end{align}
Solving \eqref{eq:opt} is challenging due to the rank constraints. As mentioned above, we approach this problem using ADMM. To that end,  \(2 m \) auxiliary matrix variables of size $9 \times 9$ are added: $m$ variables duplicating $\{F_{\tau(k)}\}_{k=1}^m$, denoted \(\{B_k\}_{k=1}^{m}\), and $m$ Lagrange multipliers $\{\Gamma_k\}_{k=1}^{m}$, yielding the objective
\begin{align} \label{eq:admm}
\max_{\Gamma_{k}} ~& \min_{F, B_1,...,B_m}
\sum_{k=1}^m L(F_{\tau(k)},B_k,\Gamma_k) & \\ 
& \mathrm{s.t.} ~~~
F=F^T \nonumber \\
& ~~~~~~~~ F_{ii}=0_{3 \times 3} & i=1,...,n \nonumber \\
& ~~~~~~~~ rank(B_{k})=6 & k=1,...,m, \nonumber
\end{align}
where
\begin{equation*}
L(F_{\tau(k)},B_k,\Gamma_k)=\alpha||\hat F_{\tau(k)}-F_{\tau(k)}||^2_F+||B_k-F_{\tau(k)}+\Gamma_k||_F^2.
\end{equation*}
We initialize the auxiliary variables at $t=0$ with  
$$ B_k^{(0)}=\hat{F}_{\tau(k)}, ~~\Gamma_{k}^{(0)}=0$$ and then alternate between the following three steps, where at each step we update the values of the variables at iteration $t$ given their values at $t-1$.

\medskip

\noindent (i) \textbf{\uline{Solving for $F$}}. \quad \begin{align} \label{eq:Fsolve}
&\underset{F}{\text{argmin}}
& & \sum_{k=1}^m \alpha||\hat F_{\tau(k)}-F_{\tau(k)}||^2_F\\
& & & \ \ \ \ \ \ \ \ \ \ \ +||B_k^{(t-1)}-F_{\tau(k)}+\Gamma_k^{(t-1)}||_F^2 \nonumber
\\ \nonumber
& \text{s. t.}
& & F=F^T \\ \nonumber
& & & F_{ii}=0_{3 \times 3}~~~~~~~ i=1,...,n. \nonumber
\end{align}
In practice, we explicitly maintain the equality constraints over $F$, i.e.,  $F$ is symmetric with zero block diagonal. Therefore, at each iteration $t$ we can solve only for the triangular upper part of $F$, i.e., 
 $\{F_{ij}| {\hat F}_{ij} \in \Omega, i < j \}$. This yields an unconstrained convex quadratic objective in these  variables, and hence it admits a closed form solution. Let $N_{ij}$ be the number of 3-cliques in $\tau$  that include the edge $e_{ij}$.  Then,  for each such triplet $\tau(k)$ we denote the variables corresponding to the $i,j$ block as $B_k(i,j)$, $\Gamma_k(i,j)$, and ${\hat F}_{\tau(k)}(i,j)$. This yields the following update rule
\begin{align}\label{eq:closedform}
& F_{ij}^{(t)}=
& \frac{1}{N_{ij}(1+\alpha)} \sum_{k=1}^{N_{ij}} B_k^{(t-1)}(i,j)+\Gamma_k^{(t-1)}(i,j)\\
& & + \alpha \hat F _{\tau(k)}(i,j).
\nonumber  
\end{align}

\medskip

\noindent (ii) \textbf{\uline{Solving for $B_k$}}. \quad For all $k=1,\ldots,m $
\begin{align}
B_k^{(t)} &= \underset{B_k}{\text{argmin}}
  ||B_k-F_{\tau(k)}^{(t)}+\Gamma_k^{(t-1)}||_F^2 \\
\text{s.t}~~~&rank(B_{k})=6.
\nonumber
\end{align}
Here, the closed form solution is
\begin{equation}
\label{eq:updateB}
B_k^{(t)}=SVP(F_{\tau(k)}^{(t)}-\Gamma_k^{(t-1)},6)
\end{equation}
where $SVP(A,p)$ denotes the singular value projection of the matrix $A$ to rank $p$.

\medskip

\noindent (iii) \textbf{\uline{Updating $\Gamma_k$}}. \quad For all $k=1,\ldots,m $
\begin{equation}
\label{eq:updateGamma}
\Gamma_{k}^{(t)}=\Gamma_{k}^{(t-1)}+B_k^{(t)}-F_{\tau(k)}^{(t)}.
\end{equation}
Note that the constraints in~\eqref{eq:opt}  cover only a subset of the constraints in Theorem \ref{thm:consistent_direct}, and in particular they do not restrict the sign pattern of the eigenvalues of $F_{\tau(k)}$, the rank 2 of $F_{ij}$, or the rank 3 of $F_i$. Our experiments, however, indicate that with the amounts of noise prevalent in existing datasets, and by removing collinear triplets, our solutions always satisfy these constraints to a good numerical precision.


%% file: method2.tex
\subsection{Camera recovery}\label{section::camera_recovery}
We use the optimized matrix $F$ to recover the corresponding projective cameras. Since $F$ is generally inconsistent we cannot use the eigen-decomposition method described in Corollary~\ref{corollary:projective_reconstruction}. However, the $9 \times 9$ sub-matrices $F_{\tau(k)}$ are consistent, and by construction they form a triplet cover of $G$. It is therefore straightforward to traverse the graph $G_\tau$ and, using Theorem~\ref{thm:merging},  apply a homography to each three cameras corresponding to $F_{\tau(k)}$, $k=1,...,m$, to bring them all to a common projective frame. Note that this process is exact, since all $F_{\tau(k)}$ are consistent.

\subsection{Constructing triangle cover}  \label{section::graph_construction}
\newcommand{\vc}[1]{\bold #1}
Eq.~\eqref{eq:opt} enforces the consistency of 3-view sub-matrices. In principle, this formulation can be applied to all 3-cliques of $G$ (although outlier and collinear triplet removal may be needed). It is however more efficient (and suffices for reconstruction) to apply this to a subset of the triplets, provided the triplets produce a solvable viewing graph.

Given a viewing graph $G=(V,E)$, 
we consider further a graph $G_\tau=(V_{\tau},E_{\tau})$ whose nodes $v' \in V_\tau$ represent 3-clique in $G$ and an edge $e'_{kl} \in E_\tau$ exists if triangles $v'_k$ and $v'_l$ share two images. We call $G_\tau$ a \textit{triplet cover} of $G$ if every node $v_i \in V$ belongs to at least one vertex in $V_\tau$.

Our objective is to find a small and reliable triplet cover of $G$. We do this heuristically as follows. We associate a weight $w_{ij}$ with each edge $e_{ij} \in E$, where \(w_{ij}\) counts the number of inliers of pairwise correspondences,  identified for $\hat{F}_{ij}$. We then find \(N_{G}\)  edge-disjoint maximal spanning trees for $G$, and use this set to produce a triplet cover for $G$, denoted $G_\tau$, in a similar way to \cite{klopschitz2010robust}. Next, we prune $G_\tau$ greedily, removing triplets whose cameras are collinear and triplets whose consistency scores are low. To measure collinearity we divide the distance between the two epipoles in each image by their average distance from the image center and average these ratios over the three images. Denote this measure by $l_k$ we remove triplets with $l_k < \delta_1$. We further measure consistency as  $c_k=\|F_{\tau (k)}-\hat{F}_{\tau (k)}\|_F$, where $\hat{F}_{\tau(k)}$ is the measured 3-view fundamental matrix associated with the triplet $v'_k$ and  $F_{\tau(k)}$  is  its  closest consistent triplet  calculated using our ADMM optimization for only this triplet. (This can be done very efficiently.) Finally, we sort the remaining triplets by their stability scores, defined as $s_k=l_k^{\delta_2} / c_k$, and greedily remove triplets of low score while maintaining the connectivity of $G_\tau$ and its cover of $G$, see an illustration in Fig.~\ref{fig:triplet}.

%% file: experiments.tex
\begin{table*}[tbh]
\tiny
\captionof{table}{\small Reprojection error and run time obtained in our experiments.}\label{table:mainresult}
\resizebox{\linewidth}{!}{%
\
\begin{tabular}{|l|r|r|llll|llll|}
\hline
 \multirow{2}{5em}{\textbf{Dataset}} & \multirow{2}{3em} {\textbf{\#points}} &\multirow{2}{3em}{\textbf{\#Images}} &\multicolumn{4}{|c|}{ \textbf{Error(pixels)} }  & \multicolumn{4}{|c|}{\textbf{Time(s)} }  \tabularnewline
\cline{4-11}
 &  &  & \textbf{Ours} & \textbf{PPSFM} & \textbf{\textbf{Sengupta}}  & \textbf{Var-Pro} &\textbf{Ours} & \textbf{PPSFM} &\textbf{Sengupta}  & \textbf{Var-Pro}\tabularnewline
\hline
Dino 319 & 319 & 36 & \textbf{0.4314} & 0.5042 & 0.6134 &  0.6157 & \textbf{3.07} & 3.24 & 46.40 & 5.48\tabularnewline
Dino 4983 & 4983 & 36 & \textbf{0.4205} & 0.4442 & 0.5795 & 0.5961 &\textbf{4.80} & 15.19  & 50.35 & 384.01\tabularnewline
Corridor & 737 & 11 & \textbf{0.2596} & 0.276 & 0.2765 & 0.2741 & \textbf{1.12} & 1.31 & 21.05 & 45.16\tabularnewline
House & 672 & 10 & \textbf{0.3399} & 0.3687 & 0.5984 & 0.3719 & 0.95& \textbf{0.70} & 18.10 & 55.19\tabularnewline
Gustav Vasa & 4249 & 18 & \textbf{0.1564} & 0.1687 & 0.2591 & 0.1671 &  \textbf{3.18} & 7.06 & 32.93 & 326.48\tabularnewline
Folke Filbyter & 21150 & 40 & \textbf{0.258} & -- & -- & 26.6054 & \textbf{6.07} & -- & -- & 2.33E+04\tabularnewline
Park Gate & 9099 & 34 & \textbf{0.3109} & 0.3447 & 0.3288 & 0.5489 & \textbf{11.33} & 25.85 & 62.6 & 1600\tabularnewline
Nijo & 7348 & 19 & \textbf{0.3901} & 0.4412 & 0.4173 & 0.417 & \textbf{5.70}& 10.33 & 32.22 & 148.74\tabularnewline
Drinking Fountain & 5302 & 14 & \textbf{0.2806} & 0.3125 & 0.2942 & 0.2942 & \textbf{2.85} & 5.80 & 26.35 & 82.07\tabularnewline
Golden Statue & 39989 & 18 & \textbf{0.223} & 0.24 & 0.2368 & 0.2272 & \textbf{7.07} & 20.45 & 86.8 & 3890\tabularnewline
Jonas Ahls & 2021 & 40 & \textbf{0.1845} & 0.2108 & 0.1979 & 0.197 & \textbf{5.29} & 15.22 & 46.64 & 90.81\tabularnewline
De Guerre & 13477 & 35 & \textbf{0.2609} & 0.2891 & 0.2728 & 0.2715 & \textbf{13.79} & 27.92 & 103.70 & 282.87\tabularnewline
Dome & 84792 & 85 & \textbf{0.2354} & 0.2507 & 1.8991 & 0.2413 & \textbf{109.83} & 171.83 & 970 & 3.78E+04\tabularnewline
Alcatraz Courtyard & 23674 & 133 & \textbf{0.5162} & 0.5592 & 5.3641 & 0.5366 & \textbf{65.04} & 113.17 & 537 & 3210\tabularnewline
Alcatraz Water Tower & 14828 & 172 & \textbf{0.4704} & 0.5972 & 0.5003 & 3.0353 & 87.11 & \textbf{68.31} & 539 & 1080\tabularnewline
Cherub & 72784 & 65 & \textbf{0.7408} & 0.7921 & -- & Out of memory (16GB)  &\textbf{34.87} & 81.42 & -- & --\tabularnewline
Pumpkin & 69335 & 195 & \textbf{0.5959} & -- &--  &  Time Limit (12H)& \textbf{130.74} & -- & -- & --\tabularnewline
Sphinx & 32668 & 70 & \textbf{0.3366}& 0.3669 & 0.3508 & 0.3486 & \textbf{23.36} &49.05 & 191 & 2.53E+04\tabularnewline
Toronto University & 7087 & 77 & 0.5417& 0.2588 & 0.2557  & \textbf{0.2556} & \textbf{25.65} & 100.27 & 779.5 & 335.9708\tabularnewline

Sri Thendayuthapani & 88849 & 98 & 0.6113 & 0.3517 & \textbf{0.3204} & Out of memory (16GB) & \textbf{248.72} & 418.21 & 1070 & --\tabularnewline
Porta san Donato & 25490 & 141 & \textbf{0.3992} &0.4352 & 4.5186 & 0.4155 & \textbf{78.22} & 87.84 & 771 & 2060\tabularnewline
Buddah Tooth & 27920 & 162 & \textbf{0.5957} & 0.8583 & 1.7853 & 0.6245 & \textbf{62.17} & 84.44 & 792 & 5530\tabularnewline
Tsar Nikolai I & 37857 & 98 & \textbf{0.2897} & 0.309 & 0.3021 & 0.3013 & \textbf{62.71} & 95.68 & 422 & 8700\tabularnewline
Smolny Cathedral & 51115 & 131 & \textbf{0.4639} & 0.5079 & 0.4773 & Out of memory (16GB) & \textbf{197.72} & 264.04 & 640.87 & --\tabularnewline
Skansen Kronan & 28371 & 131 & 0.4424 & 0.4414 & 0.4477 & \textbf{0.4291} &\textbf{102.75} &165.77 & 1000 & 2610\tabularnewline
\hline
\end{tabular}
}
\end{table*}
\begin{algorithm}
    \SetKwInOut{Input}{Input}
    \SetKwInOut{Output}{Output}
    \Input{Fundamental matrices $\Omega=\{\hat{F}_{ij}\}$\\
    Viewing Graph: $G=(V,E)$\\
    Tracks of point matches (for BA)}
    \Output{Projective reconstruction of Cameras and Points}
    $G_\tau \leftarrow $Select a triplet cover for $G$ (Sec.~\ref{section::graph_construction})\\
    Form the $n$-view measurement matrix $\hat{F}\leftarrow \{\hat{F}_{ij}\}$\\
    \textbf{\uline{Solve \eqref{eq:opt} using ADMM}:}\\
    Initialize $B_k^{(0)}=\hat{F}_{\tau(k)} , ~~\Gamma_{k}^{(0)}=0,t=1$\\
    \For{$t = 1, ..., N_{it}$ }{
Update $F^{(t)}$ using \eqref{eq:closedform}\\
For $k=1,...,m,$ update $B_k^{(t)}$ using \eqref{eq:updateB}\\
For $k=1,...,m,$ update $\Gamma_k^{(t)}$ using \eqref{eq:updateGamma}}
For $k=1,...,m$, retrieve camera matrices for triplet $\tau(k)$ (Corollary~\ref{corollary:projective_reconstruction})\\ 
    $\cal{P} \leftarrow$ Bring cameras to a global projective frame of reference (Sec. \ref{section::camera_recovery})\\
    $\cal{X} \leftarrow$ Triangulate points from points tracks\\
$\cal{P,X}\leftarrow $ Refine solution  using Bundle Adjustment\\
\textbf{Return} $\cal{P,X}$ 
\caption{ Projective SFM Pipeline}
\label{Alg:alg2}
\end{algorithm}

\begin{figure}[tb]
\centering
\includegraphics[width=0.32\linewidth]{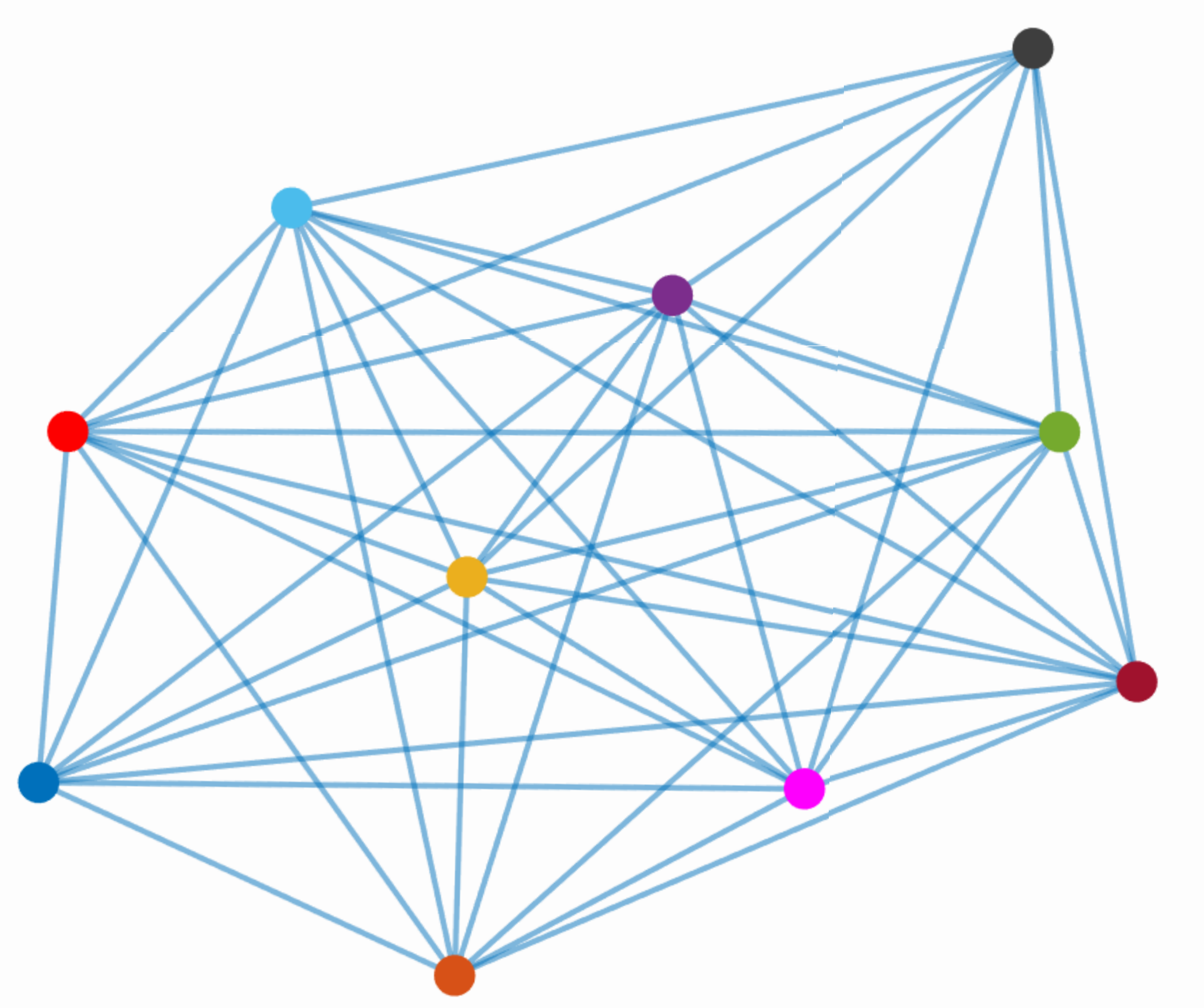}\ 
\includegraphics[width=0.32\linewidth]{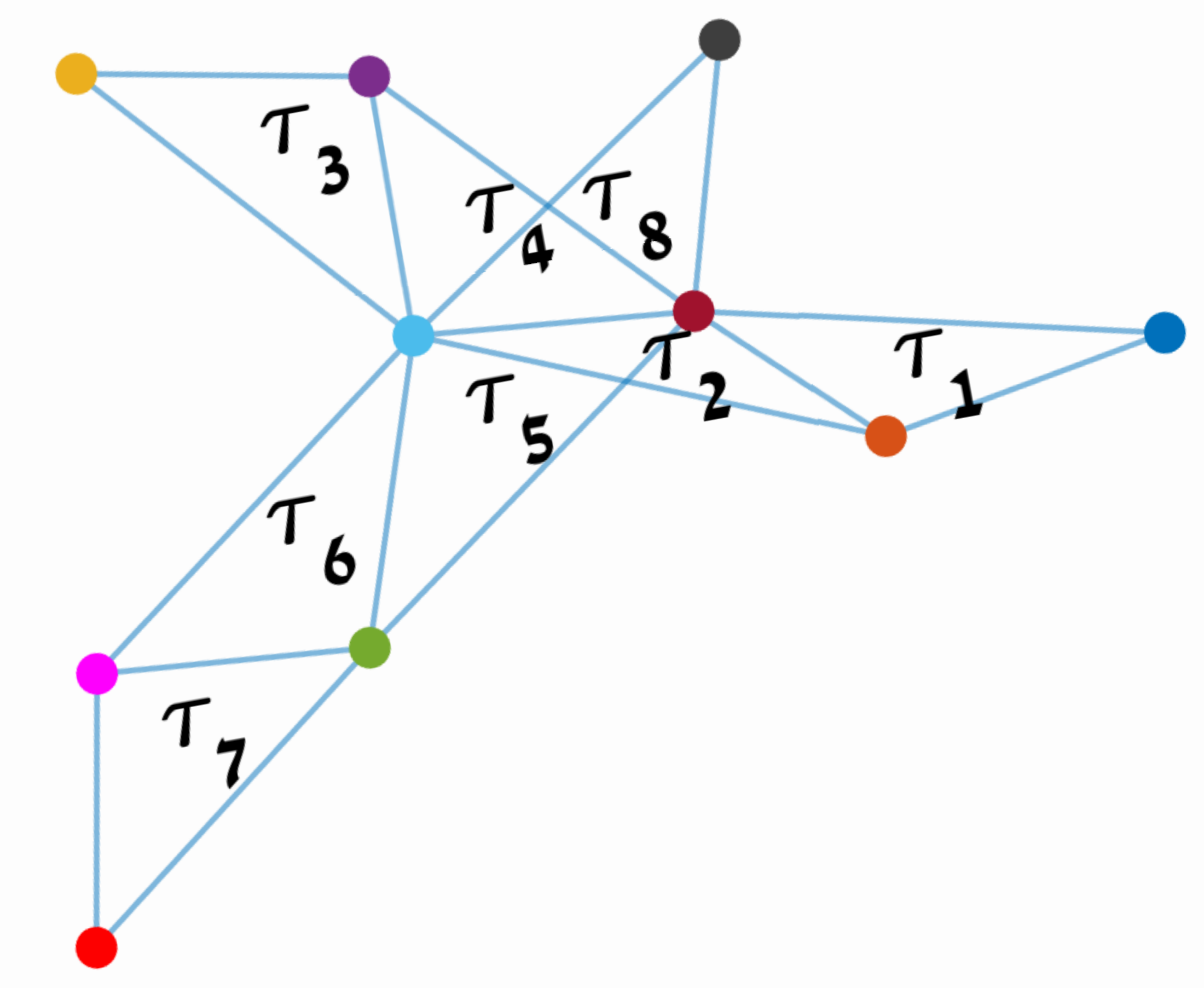}\
\includegraphics[width=0.32\linewidth]{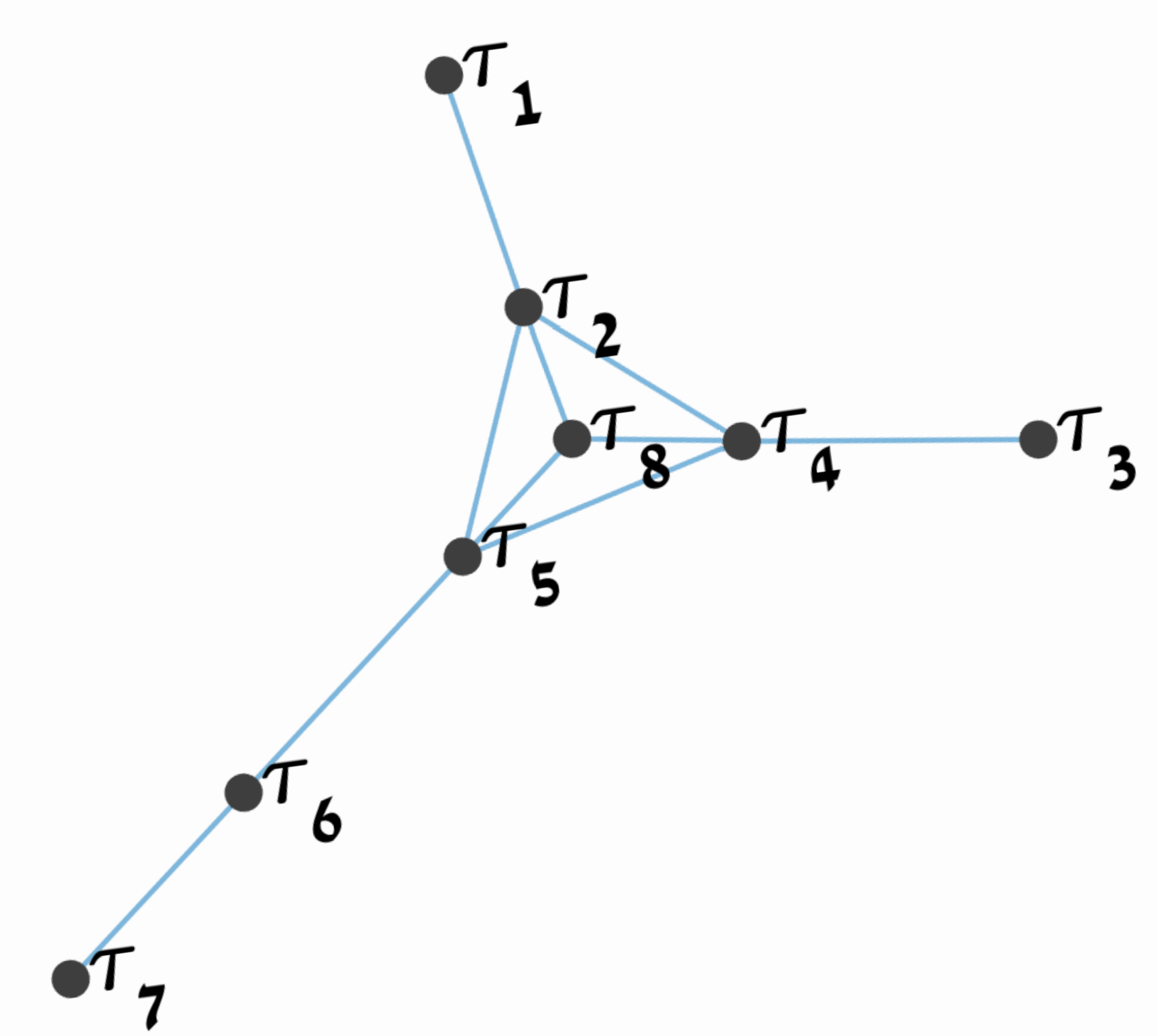}
\caption{\small
Building a triplet cover for the House dataset (10 cameras). From left to right, the viewing graph, the final viewing graph, and the corresponding triplet cover.}
\label{fig:triplet}
\end{figure}

\subsection{Structure from motion pipeline }
We used our method to construct a projective SFM pipeline. The pipeline obtains pairwise fundamental matrices computed with RANSAC. As described in Sec.~\ref{section::method}, we first construct a reliable triangle cover. We next use ADMM to solve \eqref{eq:opt}, obtaining consistent 3-view submatrices,  and use them to construct camera matrices. In postprocessing we use projective bundle adjustment (BA) to improve our camera recovery and 3D point reconstruction. Note that, similar to global Euclidean SFM methods \cite{arie2012global,goldstein2016shapefit,jiang2013global,ozyesil2015robust,wilson2014robust}, we only apply BA once at the end of the pipeline. For this step we triangulate point tracks and apply projective bundle adjustment to the camera matrices and the point matches. Similar to \cite{schonberger2016structure}, after convergence, 3D points are re-triangulated and a few additional iterations of BA are applied for better convergence. Our pipeline is summarized in Alg. \ref{Alg:alg2}.
\subsection{Implementation details }
Before optimization we normalize the input fundamental matrices $\hat{F}_{ij}$, as in \cite{hartley1995defence}, by $\hat F^n_{ij} = N_i^{-T}\hat F_{ij} N_j^{-1}$, where $N_i \in \Real^{3 \times 3}$ normalizes the location of interest points in image \(I_{i}\) to have zero mean and unit variance. In data sets where the point matches distribute non-isotropically, we normalize the variance separately in each axis. After the optimization $F_{ij}$ is denormalized by $N_i^TF_{ij}N_j.$ 

For the optimization, we set $\alpha=0.001$ (in \eqref{eq:Fsolve}) and perform $N_{it}=1000$ iterations of  ADMM updates. For the triplet selection procedure (Sec. \ref{section::graph_construction}) we set  \(N_{G}=5\) and $\delta_1=0.03.$
We set $\delta_2$ by the following condition, if the average non-collinearity measure, $l_k$, exceeds 0.5 we set $\delta_2 =0$. Otherwise, it means the data is highly collinear, and so  we set $\delta_2=1.2$.

To produce 3D points from multiview tracks we used the Matlab linear triangulation code of \cite{hartley2003multiple}.  We implemented projective bundle adjustment using the Ceres \cite{ceres-solver} non linear least squares optimization package and used Huber loss (with parameter 0.1) for robustness. We performed up to 100 iterations of bundle adjustment.
 Our code is implemented in Matlab on an Intel processor i-7 7700 with 16GB RAM.  

\subsection{Results} 

We tested the method on several projective datasets from \cite{olsson2011stable} and VGG \cite{vgg_dataset} and compared the results to state-of-the-art projective reconstruction pipelines, including:

\noindent \textbf{P$^2$SfM} \cite{magerand2017practical}. This recent method solves for projective structure  incrementally by solving linear least squares systems that incorporate constraints on the sought projective depths. The paper  demonstrated both superior re-projection accuracy and running time, compared to existing methods.

\noindent \textbf{VarPro} \cite{hong2016projective}. This method first applies affine bundle adjustment followed by projective BA. It further uses variable projection to improve the solution of BA.

\noindent \textbf{Sengupta et.\ al}~\cite{sengupta2017new}. Similar to our method, this method applies rank constraints to $n$-view matrices, but it explicitly recovers the scale factors. As the authors acknowledge, their algorithm is sensitive, and so it was suggested as a refinement to a calibrated method \cite{ozyesil2015robust}. To avoid calibration, for a fair comparison, we initialized the method with P$^2$SfM \cite{magerand2017practical}. Moreover, as the method does not suggest a 
way to produce projective reconstruction, we used Corollary \ref{corollary:projective_reconstruction} to obtain camera matrices from its output.

To compare all the methods, we ran them with the code supplied by the authors.   For \cite{sengupta2017new} we counted only its running time (excluding the running time of the initialization method). For \cite{hong2016projective} we set a time limit of 12 hours. We ran all the methods on the same computer under the same conditions.

\begin{figure}[b]
\centering
\subfloat{\includegraphics[width=0.3\linewidth]{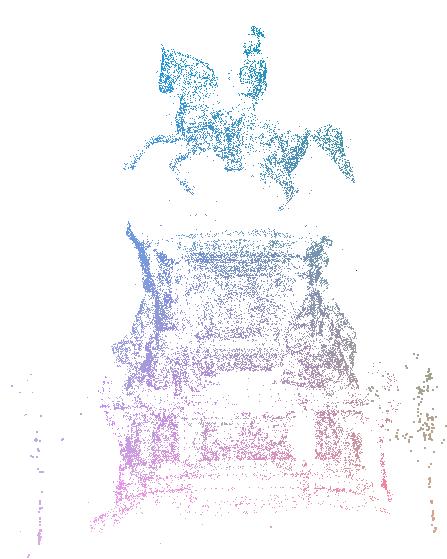}}
\subfloat{\includegraphics[width=0.3\linewidth]{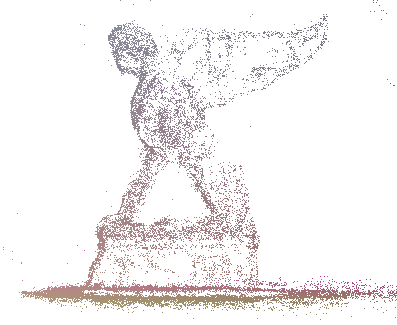}}
\subfloat{\includegraphics[width=0.3\linewidth]{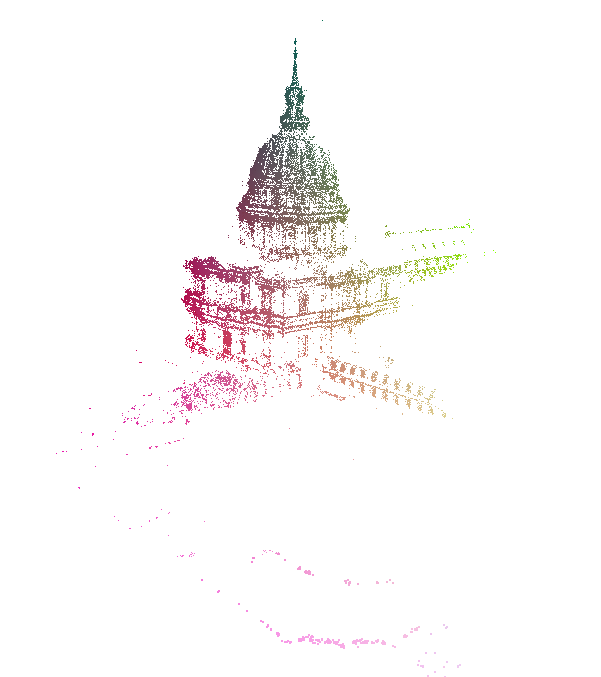}}
\caption{ \small 
Visualization of the result of our projective structure from motion pipeline after applying self-calibration of \cite{chandraker2007autocalibration} . From left to right: Tsar Nikolai I, Sphinx, Dome.  }\label{figure:pointClouds}
\label{figure:point_cloud}
\end{figure}

Table~\ref{table:mainresult} shows the mean reprojection error (in pixels) across all scene points and the total running time (in seconds)  obtained with our method compared to P$^2$SfM, VarPro, and Sengupta et al. It can be seen that our method achieved superior accuracies to all the other methods in 22 of the 25 data sets tested. Moreover, for certain complex scene structures (e.g., Pumpkin and Folk Filbyter) our method managed to reconstruct the scene, whereas all the other methods failed to obtain a reconstruction (we tried with many different hyper-parameters). In almost all cases our method was also faster, improving runtime in 22 out of 25 data sets. The best compared method (separately for each dataset) had a median of 4.8\%\ reprojection error  worse than our method and required an additional median runtime of 74\% compared to our method. The quality of our reconstruction can also be appreciated by the recovered 3D point clouds shown in Fig~\ref{figure:point_cloud}.

\begin{figure}[tb]
\centering
\includegraphics[width=0.4\linewidth]{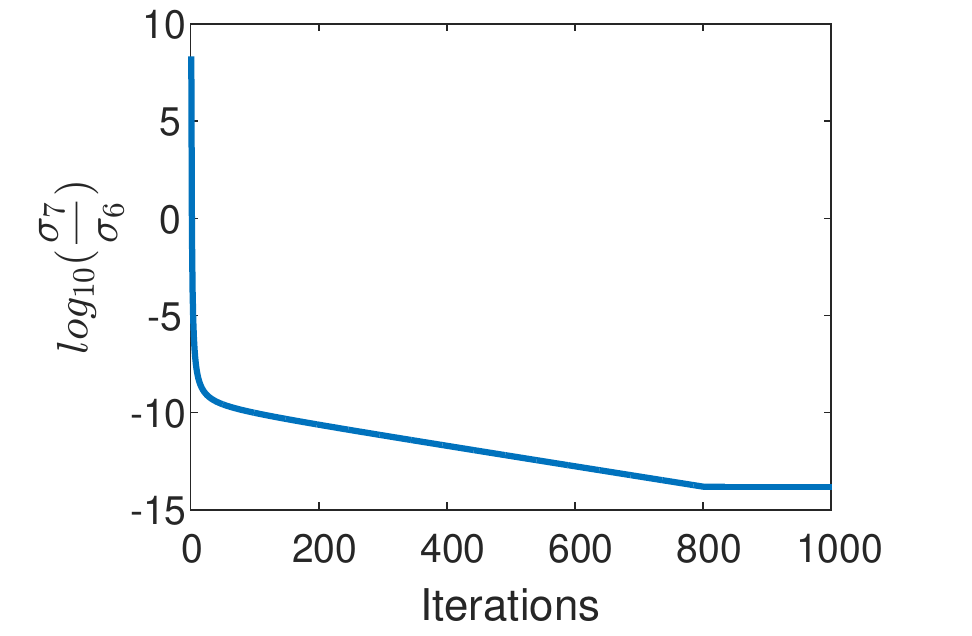}
\caption{\small Enforcing rank constraints. The plot shows (in $\text{log}_{10}$ scale) the ratio between the seventh and sixth singular value of $F_{\tau(k)}$ averaged for all triplets.}
\label{fig:sig7}
\end{figure}

We believe the improved accuracy is partly explained by the effective enforcement of the rank 6 constraint for all triplets. We demonstrate this in Fig.~\ref{fig:sig7}, which shows the ratio between the 7$^{th}$ and 6$^{th}$ singular values of $F_{\tau(k)}$ averaged over all triplets for the House model. Our optimization reduces this ratio to near machine precision, indicating that indeed rank 6 is achieved in all runs.

\subsection{Graph consistency simulations}

Since our optimization \eqref{eq:opt} only enforces a subset of the consistency constraints in Theorem \ref{thm:consistent_direct} we next test the consistency of our recovered fundamental matrices (with no bundle adjustment) in synthetic experiments. We generated 10 camera matrices and 15,000 three dimensional points. We then projected the points and perturbed them by Gaussian noise. Finally, we selected 15 matching triplets at random and used them to compute fundamental matrices, which we gave to our
algorithm. 

\begin{figure}[tb]
\centering
\subfloat[]{\includegraphics[width=0.33\linewidth]{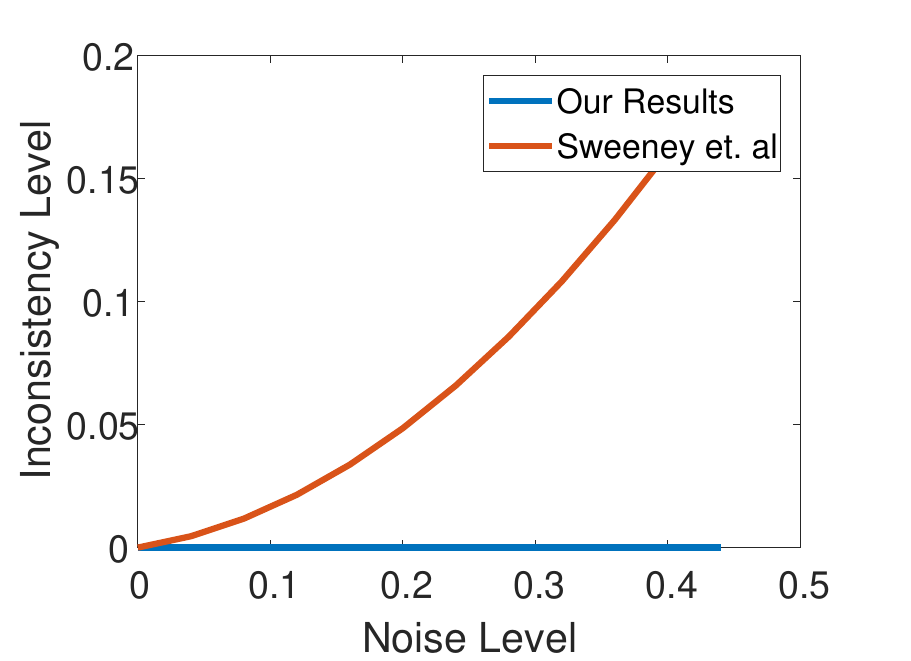}}
\subfloat[]{\includegraphics[width=0.33\linewidth]{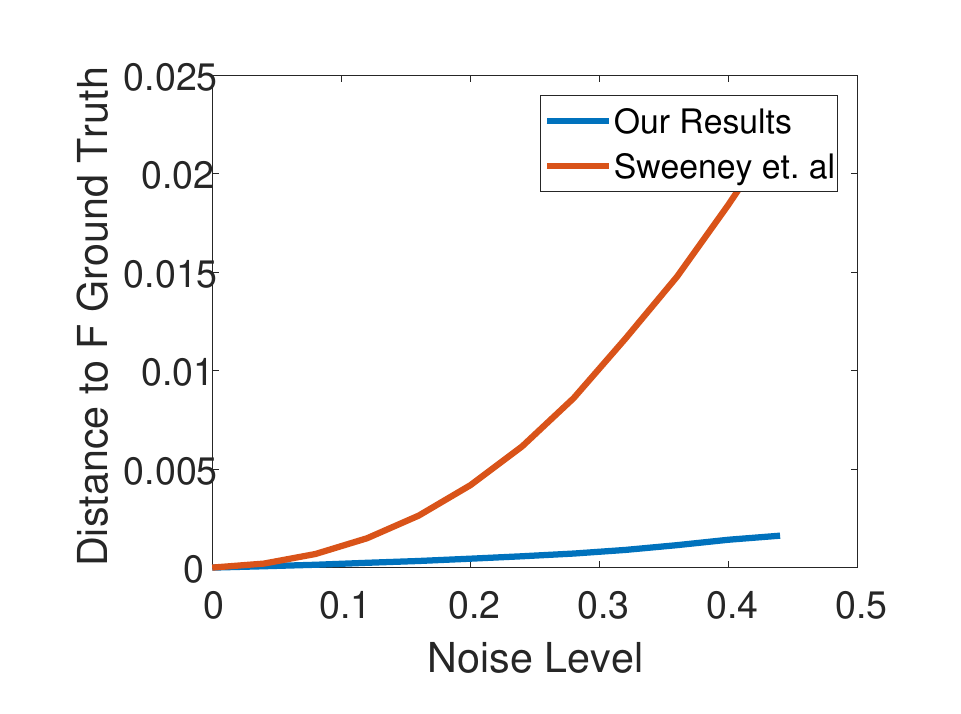}}
\subfloat[]{\includegraphics[width=0.33\linewidth]{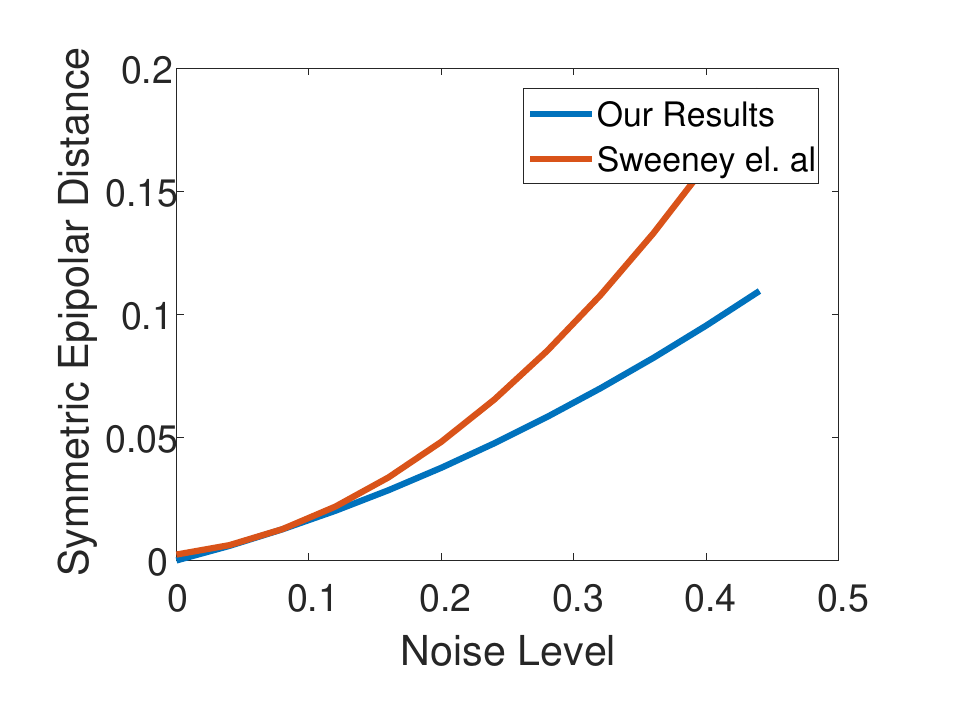}}
\caption{\small Synthetic experiments. Average symmetric epipolar distance of corresponding epipoles (left), error in fundamental matrix recovery, compared to ground truth (average Frobenius norm, middle), and symmetric epipolar distance for ground truth matches. Our method (in blue) is compared against 
\cite{sweeney2015optimizing}.}
\label{Consistencygraph}
\label{figure:sweeney_ours}

\end{figure}

The results are shown in Fig.~\ref{Consistencygraph}. We evaluated consistency using the symmetric epipolar distance associated with the term $\mathbf{e}_{ik}^{T}F_{ij}\mathbf{e}_{jk}$, where $\mathbf{e}_{ik}$ denotes the projected location of camera $k$ onto
camera $i$. Denote this distance by $S_{ijk}$, then $S_{ijk}+S_{jki}+S_{kij}=0$ implies that cameras $i,j,k$ are consistent (\cite{hartley2003multiple}, p. 384). Additionally, we show the quality of recovering the ground truth fundamental matrices (measured by average Frobenius norm), and the average symmetric epipolar distance of the ground truth matches. It can be seen that our method maintains consistency under all error levels while achieving high quality recovery of fundamental matrices compared to ground truth. We compare our results with  \cite{sweeney2015optimizing}'s consistency optimization (Sec. 3 therein).  

\section{Conclusion}

We considered the problem of recovering projective camera matrices from collections of fundamental matrices. We derived a complete algebraic characterization of $n$-view fundamental matrices in the form of conditions that are both necessary and sufficient to enable the recovery of camera matrices. We further introduced an algorithm that uses this characterization for global recovery of camera matrices from measured fundamental matrices. Our algorithm is efficient and requires no initialization. We tested the algorithm on a large number of datasets and compared it to existing, state-of-the-art methods, showing both improved accuracy and runtime.

In future work we plan to explore ways to relate our algebraic constraints with the question of solvability of viewing graphs. In addition, we plan to similarly characterize collections of essential matrices.

{\small \noindent\textbf{Acknowledgment} This research was supported in part by the Minerva foundation with funding from the Federal German Ministry for Education and Research.}

%% file: SM.tex
\appendix
\section*{Appendix}

\title{GPSfM: Global Projective SFM Using Algebraic Constraints\\ on Multi-View Fundamental Matrices \\
-Supplementary Material-}
\author{Yoni Kasten* \hspace{1cm} Amnon Geifman* \hspace{1cm} Meirav Galun  \hspace{1cm}Ronen Basri \\
Weizmann Institute of Science\\
{\tt\small  \{yoni.kasten,amnon.geifman,meirav.galun,ronen.basri\}@weizmann.ac.il}
}



Below we prove Theorem 1. For clarity, we base the proof on several supporting lemmas, whose proofs  follow. Also, our proof of Theorem 1 relies partly on necessary conditions that were introduced and proved in \cite{sengupta2017new}. Those conditions are summarized in Lemma \ref{lemma:consistent}.       
\makeatletter
\def\blfootnote{\xdef\@thefnmark{}\@footnotetext}
\makeatother
\blfootnote{*Equal contributors}
\input{theoretical_partSM}

\input{LemmasSM}





%% file: theoretical_partSM.tex
\setcounter{theorem}{0}
\begin{theorem}\label{thm:consistent_directSM} 

An n-view fundamental matrix $F$ is consistent with a set of $n$ cameras whose centers are not all collinear if, and only if, the following conditions hold: 
\begin{enumerate}
\item $Rank(F)=6$ and $F$ has exactly 3 positive and 3 negative eigenvalues.
\item $Rank(F_i) = 3$ for all $i =1, ..., n$.
\end{enumerate}
\end{theorem}

\begin{proof}
\noindent The proof of the necessary conditions relies the properties of symmetric matrices specified in Lemma \ref{lemma:eigenvalues} and on \cite{sengupta2017new}, whose derivations are summarized in Lemma \ref{lemma:consistent}.
Specifically, let $F$ be a consistent, $n$-view fundamental matrix. Then, according to Lemma \ref{lemma:consistent}, $F$ can be  written as $F=UV^T+VU^T$, where $U,V \in \Real^{3n \times 3}$ whose $3 \times 3$ blocks respectively are $U_i=V_iT_i$ and $V_i$. And, moreover, since the camera centers are not all collinear, we have $rank(F) =6$, $rank(U)=3$ and $rank(V)=3$, implying property (\textit{iii}) of Lemma \ref{lemma:eigenvalues}. Consequently, using property (\textit{i}) of  Lemma \ref{lemma:eigenvalues}, condition 1 holds. Condition 2 holds because not all cameras are collinear, since if conversely  $rank(F_i) < 3$ for some $i$ then there exists a 3-vector ${\bf e} \neq 0$ such that $F_i^T {\bf e}=0$, and therefore $\forall j~F_{ji} {\bf e}=0$, i.e., all epipoles collapse to the same point in frame $i$, implying, in contradiction, that the   camera centers are all collinear. 

To establish the sufficient condition, let $F$ be an $n$-view fundamental matrix that satisfies conditions 1 and 2. Condition 1 (along with property (\textit{iii}) of  Lemma \ref{lemma:eigenvalues}) implies that $F$ can be decomposed into $F=U V^T + V U^T$. This decomposition, along with condition 2, allows to deduce, using Lemma \ref{lemma:xor}, that WLOG $\forall i, rank(V_{i})=3$ and $rank(U_{i})=2$. This, and the skew-symmetry of $U_iV_i^T$ (due to $F_{ii}=0$), imply, using Lemma \ref{lemma:skew}, that $V_i^{-1}U_i$ is skew-symmetric. Denote this matrix by $T_i=[{\bf t}_i]_{\times}$, we obtain $F_{ij}=V_{i}(T_{i}-T_{j})V_{j}^{T}$, establishing that $F$ is consistent. Finally, $\{{\bf t}_i\}_{i=1}^n$ are not all collinear, since, otherwise, by Lemma \ref{lemma:collinear}, $\exists i$ and $\exists {\bf e} \neq 0$ such that $\forall j~F_{ji} {\bf e}=0$, implying that $F_i^T {\bf e}=0$, contradicting the full rank of $F_i$.

\end{proof}

%% file: LemmasSM.tex
\setcounter{theorem}{0}

\label{section::sufficient_conditions}

We next turn to stating and proving the supporting lemmas.

\begin{lemma}\label{lemma:eigenvalues}
Let $F \in \mathbb{S}\ ^{3n}$ be a matrix of rank 6. Then, the following three conditions are equivalent. 
\begin{enumerate}
\item[(i)] $F$ has exactly 3 positive and 3 negative eigenvalues.
\item[(ii)] $F=XX^T-YY^T$ with $X,Y \in \Real^{3n \times 3}$ and $rank(X)=rank(Y)=3.$
\item[(iii)] $F=UV^T+VU^T$ with $U,V \in \Real^{3n \times 3}$ and $rank(U)=rank(V)=3$.
\end{enumerate}
\end{lemma}
\begin{proof}
Assume (\textit{i}), and denote the eigenvalues of $F$
by $\lambda_1 \geq \lambda_2 \geq \lambda_3 > 0 > \lambda_4 \geq \lambda_5 \geq \lambda_6$. Applying spectral decomposition to $F$ we obtain
\begin{align*}
F & =[\tilde{X},\tilde{Y}]\left(\begin{array}{cc}
\Sigma_{1} & 0\\
0 & -\Sigma_{2}
\end{array}\right)[\tilde{X},\tilde{Y}]^{T} \\
&=\tilde{X}\Sigma_{1}\tilde{X}^{T}-\tilde{Y}\Sigma_{2}\tilde{Y}^{T},
\end{align*}
where $\tilde{X}, \tilde{Y} \in \Real^{3n \times 3}$, $\Sigma_1 = \text{diag}(\lambda_1, \lambda_2, \lambda_3)$ and $\Sigma_2 = \text{diag}(-\lambda_4,- \lambda_5,- \lambda_6)$.  
Next, we define $X = \tilde{X} \sqrt{\Sigma_1}$ and $Y = \tilde{Y} \sqrt{\Sigma_2}$ then 
\begin{align*}\label{eq:decomposition}
F & = XX^{T}-YY^{T},
\end{align*}
where $rank(X)=rank(Y)=3$, implying (\textit{ii}). Next, let $U=\sqrt{\frac{1}{2}}(X+Y)$ and $V=\sqrt{\frac{1}{2}}(X-Y)$. It can be readily verified that
\begin{equation*}
F=UV^{T}+VU^{T}.
\end{equation*}
Moreover, if either $U$ or $V$ are rank deficient then $rank(F) < 6$, contradicting the assumption. Therefore, $rank(U) = rank(V)=3$, implying (\textit{iii}).
 
To complete the proof, assume (\textit{iii}), i.e.,  $F=UV^{T}+VU^{T}$, where $U, V \in \Real^{3n\times3}$ are of rank 3. We define  $X=\sqrt{\frac{1}{2}}(U+V)$ and $Y=\sqrt{\frac{1}{2}}(U-V)$ yielding  $F=XX^{T}-YY^{T}$, with $rank(X) = rank(Y) =3$, implying (\textit{ii}).
 
It remains to  show that $(\textit{ii}) \Rightarrow (\textit{i})$.
Since $F$ is symmetric of degree 6, it has exactly 6 real, non-zero eigenvalues. We now show that exactly 3 of these eigenvalues are positive and 3 are negative. By contradiction,
assume w.l.o.g.\ that $F$ has at least 4  positive eigenvalues, denoted by $\lambda_{1},\lambda_{2},\lambda_{3},\lambda_{4}$, and denote their
corresponding eigenvectors by $v_{1},v_{2},v_{3},v_{4}$. Denote the subspace spanned by these eigenvectors by $S$, i.e., $S =span\{v_{1},v_{2},v_{3},v_{4}\} $. Now, due to orthogonality, for every $\sum_{i=1}^{4} a_{i}v_{i}=z\in S$
we have $$Fz = \sum_{i=1}^4 \alpha_i \lambda_i v_i  \Rightarrow z^T Fz= \sum_{i=1}^4 \alpha_i^2 \lambda_i.$$  Therefore, since $\lambda_i > 0$, for $0 \neq z \in S$ we have,  $$z^T F z = \sum_{i=1}^4 \alpha_i^2 \lambda_i > 0.$$
On the other hand, the dimension of the column space of $X$ is at most 3 and therefore  $\exists \bar z \in S$, which is orthogonal to the column space of $X$, i.e. $X^T\bar z=0$, implying that
\begin{equation*}
\bar z^T F \bar z = \bar z^T (X X^T - Y Y^T)\bar z = - \bar z^T Y Y^T \bar z \leq 0,
\end{equation*}
which contradicts our previous observation that every vector $0 \neq z\in S$
satisfies $z^{T}Fz>0$.
 The same argument can be applied to the negative
eigenvalues. We conclude  that $F$ has exactly 3 positive eigenvalues and 3 negative eigenvalues.
\end{proof}


\begin{lemma}\label{lemma:consistent} \cite{sengupta2017new}  Let $F$ be a consistent $n$-view fundamental matrix. Then, 
\begin{enumerate}
\item 
$F$ can be formulated as $F=UV^{T}+VU^{T}$, where $V,U \in \Real^{3n \times 3}$ consist of $n$ blocks of size $3 \times 3$ 
\begin{align*}
V =\left[\begin{array}{ccc}
V_{1}\\
\vdots\\
V_{n}
\end{array}\right] ~~ & ~~
U =\left[\begin{array}{c}
V_{1} T_{1}\\
\vdots\\
V_{n} T_{n}
\end{array}\right]
\end{align*}
and $T_i = [{\bf t}_i]_{\times}$.
\item $rank(V) = 3$ 
\item If ${\bf t}_i$ are not all collinear then $rank(U)=3$ and $rank(F) = 6$. 
\end{enumerate}
\end{lemma}

\begin{proof}
Condition 1 follows directly from Eq.~(1) in the paper, namely
\begin{equation*} 
F_{ij} = V_i(T_i-T_j)V_j^T.
\end{equation*}
Condition 2 is satisfied since $V_i$ is invertible for all $i=1, ..., n$. Next, we prove Condition 3 by contradiction. Assume $rank(U) < 3$. Then, $\exists {\bf t} \neq 0$, s.t. $U {\bf t}=0$. Since $V_i$ are of full rank for all $i=1, ..., n$, this implies that ${\bf t}_i \times {\bf t} = 0$ for all $i=1, ..., n$. Thus, all the ${\bf t}_i$'s are parallel to ${\bf t}$, violating our assumption that not all ${\bf t}_i$ are collinear.

We are left to show that if ${\bf t}_i$ are not all collinear then $rank(F)=6$. Using the QR decomposition for an invertible matrix, each $V_i$ can be decomposed uniquely into a product of a  lower triangular matrix with positive diagonal elements and an orthogonal matrix. Therefore, there exist  an upper triangular $K_i$ and an orthogonal matrix $R_i$ such that $V_i = K_i^{-T} R_i^T$. We can thus write $F = K^T E K$, where the $3n \times 3n$ matrix $K$ is a block diagonal matrix with $3 \times 3$ blocks formed by $\{ K_i^{-1} \}_{i=1}^n$, and so it has full rank, implying that $rank(F) = rank(E)$. We are left to show that $rank(E)=6$. Since $E$ has the same form as in \cite{sengupta2017new}, the proof can be completed as described there (\cite{sengupta2017new}, p.~3).
\end{proof}

\begin{lemma}\label{lemma:common_null_space} Let $A,B\in \Real^{3\times3}$such that $rank(A)=rank(B)=2$
and $AB^{T}=[\bf{t}]_{\times}$ for some ${\bf t}\in \Real^{3}$ then $A^{T}{\bf t}=B^{T}{\bf t}=0$
\end{lemma}

\begin{proof}
Let ${\bf t}_1 \in Ker(A^T)$ and ${\bf t}_2 \in Ker(B^T)$, ${\bf t}_1, {\bf t}_2 \ne 0$. Note also that  $AB^T = [\bf{t}]_{\times}$ implies $BA^T = - [{\bf t}]_{\times}$. Then,
\begin{align*}
A^T {\bf t}_1 = 0 & \Rightarrow BA^T {\bf t}_1 = 0 \Rightarrow {\bf -t} \times {\bf t}_1 = 0\\ & \Rightarrow {\bf t}_1 \parallel {\bf t} \Rightarrow A^T {\bf t} = 0.
\end{align*}
\begin{align*}
B^T {\bf t}_2 = 0 & \Rightarrow AB^T {\bf t}_2 = 0 \Rightarrow {\bf t} \times {\bf t}_2 = 0\\
& \Rightarrow {\bf t}_2 \parallel {\bf t} \Rightarrow B^T {\bf t} = 0
\end{align*}
\end{proof}

\begin{lemma}\label{lemma:skew} Let $A,B\in \Real^{3\times3}$ with $rank(A)=2,\,\,rank(B)=3$
and $AB^{T}$ is skew symmetric (that is $AB^{T}+BA^{T}=0$) , then
$T=B^{-1}A$ is skew symmetric.
\end{lemma}

\begin{proof} Since $AB^{T}$ is skew symmetric it can be written 
as  $AB^{T}=[\bf{a}]_{\times}$ for some ${\bf a}\in \Real^{3}\Rightarrow$
\begin{align*}
A=[{\bf a}]_{\times}B^{-T}&=BB^{-1}[{\bf a}]_{\times}B^{-T}\\&=B(B^{-1}[{\bf a}]_{\times}B^{-T})\\&=B\frac{[B^{T}{\bf a}]_{\times}}{det(B)} \end{align*}
where the last equality follows from the following identity which holds for $B \in \Real^{3 \times 3}$  $$(B {\bf x}) \times (B{\bf y}) = det(B) B^{-T} ({\bf x} \times {\bf y}).$$
Consequently, $T=B^{-1}A=\frac{[B^{T}a]_{\times}}{det(B)}$ is skew symmetric.

\end{proof}

\begin{lemma} \label{lemma:xor}
Let $F$ be an n-view fundamental matrix. If $F$ can be formulated as $F=UV^T+VU^T$ where $U, V \in \Real^{3n \times 3}$ and in addition $rank(F_i)=3$ for $i=1,\ldots,n$ then it holds that either $\forall i\,rank(V_{i})=3,\,rank(U_{i})=2$ or that $\forall i~rank(V_{i})=2,\,rank(U_{i})=3$.
\end{lemma}

\begin{proof}{
First, since $\forall i~F_{ii}=0$, it follows that $\forall i ~U_{i}V_{i}^{T}$ is skew-symmetric, and w.l.o.g we assume that $rank(U_{i}V_{i}^{T})\neq 0$ otherwise it is possible to apply Lemma \ref{lemma:gpsfm_correction_main} repeatedly to get a different decomposition that fulfills this assumption.  Therefore, $rank(U_{i}V_{i}^{T})=2$ implying that both $2 \le rank(U_{i}) \le 3$ and $2 \le rank(V_{i}) \le 3$, but both cannot have full rank}. Of the remaining possibilities. 
\begin{enumerate}

\item  $\exists i$ such that $rank(U_i) = rank(V_i)=2$. According to Lemma \ref{lemma:common_null_space}, $\exists {\bf t} \in \Real^3$, such that $U_i^T {\bf t} = V_i^T {\bf t}=0$, implying that $F_i^T {\bf t} = (V U_i^T + U V_i^T) {\bf t} = 0$. However, this contradicts the full rank assumption over $F_i$.

\item  
Suppose, without loss of generality, that 
\begin{align*}
rank(V_{1}) & =3,~rank(U_{1})=2\\
rank(V_{2}) & =2,~rank(U_{2})=3.
\end{align*}
By Lemma \ref{lemma:skew}, since $U_1 V_1^T$ is skew symmetric and $rank(V_1)=3, rank(U_1)=2$, we obtain that $T_1 = V_1^{-1}U_1$ is skew symmetric. By similar considerations $T_2 = U_2^{-1}V_2$ is skew symmetric. This yields 
\begin{align*}
F_{12} & = U_{1}V_{2}^{T}+V_{1}U_{2}^{T} \\
& =V_{1}T_{1}(-T_{2})U_{2}^{T}+V_{1}U_{2}^{T}\\
 & =V_{1}(-T_{1}T_{2}+I)U_{2}^{T}.
\end{align*}
Now, using the fact that $rank(V_1) = rank(U_2) =3$, we obtain  
\begin{equation}\label{eq:rank} rank(-T_{1}T_{2}+I) =rank(F_{12})=2. \end{equation}
In the next steps we show a contradiction to \eqref{eq:rank}.  
Since $rank(-T_{1}T_{2}+I)=2$ then $\exists {\bf v} \in null(-T_{1}T_{2}+I)$, ${\bf v} \ne 0$ for which
\begin{align*}
(-T_{1}T_{2}+I) {\bf v} = 0 & \Rightarrow T_{1}T_{2} {\bf v} = {\bf v}\\
& \Rightarrow {\bf t}_{1}\times({\bf t}_{2}\times {\bf v}) = {\bf v}. 
\end{align*}
We conclude that   ${\bf t}_1^T {\bf v} = 0$. Using the identity  
$a\times(b\times c) = b(a^{T}c)-c(a^{T}b) $, we obtain
\begin{align*} 
{\bf t}_{1}\times({\bf t}_{2}\times {\bf v}) & ={\bf v} \Rightarrow\\
{\bf t}_{2}({\bf t}_{1}^{T}{\bf v})-{\bf v}({\bf t}_{1}^{T}{\bf t}_{2}) & ={\bf v} \Rightarrow \\
-{\bf v}({\bf t}_{1}^{T}{\bf t}_{2}) & ={\bf v} \Rightarrow \\
({\bf t}_1^T {\bf t}_2) &= -1.
\end{align*} 
Now, the subspace defined by $\{ {\bf u} \in \Real^3 | {\bf t}_1^T{\bf u}=0\} $ is of dimension 2. However, as we show below, it is contained in $null(-T_{1}T_{2}+I)$,  contradicting \eqref{eq:rank}, since any vector ${\bf u}$ in this space satisfies 
\begin{align*}
(-T_{1}T_{2}+I){\bf u} & = 
-{\bf t}_1 \times ({\bf t}_2 \times {\bf u}) +{\bf u}\\ & = 
-{\bf t}_2({\bf t}_1^T {\bf u}) + {\bf u}({\bf t}_1^T {\bf t}_2) + {\bf u}\\
& = -{\bf u} + {\bf u} = 0.
\end{align*}
\end{enumerate}
Consequently, either $\forall i\,rank(V_{i})=3,\,rank(U_{i})=2$ or $\forall i\,rank(V_{i})=2,\,rank(U_{i})=3$.
\end{proof}

For the next Lemma we note that in Lemma \ref{lemma:xor} and Theorem \ref{thm:consistent_direct} we use the following property, which we justify below  $rank(F_i) = 3 \Leftrightarrow rank(F_i^T) = 3 \Leftrightarrow null(F_i^T)=\emptyset \Leftrightarrow \nexists {\bf t} \in \Real^3$, ${\bf t}\ne 0$, s.t.\ $F_i^T {\bf t} = 0 \Leftrightarrow \nexists {\bf t} \in \Real^3$, ${\bf t}\ne 0$, s.t.\ $\forall j F_{ji} {\bf t}=0$. 

\begin{lemma}\label{lemma:collinear}
Let $V_1,...,V_n \in \Real^{3 \times 3}$ and ${\bf t}_1, ..., {\bf t}_n \in \Real^3$. We define $F_{ij} = V_i[{\bf t}_i - {\bf t}_j]_{\times} V_j^T$ and assume that for $i \neq j~rank(F_{ij})=2$. Then, $\{ {\bf t}_i\}_{i=1}^n$ are collinear if and only if $\exists i \in \{1,...,n\}$ and $\exists {\bf e} \in \Real^3, {\bf e} \neq 0$, s.t.\ $\forall j F_{ji} {\bf e} = 0$. 
\end{lemma}
\begin{proof}
$\Rightarrow$ We first assume that $\{{\bf t}_i\}_{i=1}^n$ are collinear. We show it by construction. Let us choose $i \neq 1$ and define 
$$ {\bf e} = V_i^{-1}({\bf t}_i - {\bf t}_1).$$
Since ${\bf t}_i \neq {\bf t}_1$ (otherwise the rank assumption is violated) then ${\bf e} \neq 0$ and the collinear points  ${\bf t}_1,\ldots,{\bf t}_n$  can be parameterized as follows 
$${\bf t}_k = {\bf t}_1 + \alpha_k ({\bf t}_i - {\bf t}_1) ~~\forall k.$$ Now, $\forall j$ it holds that 
\begin{align*}
F_{ji} {\bf e} &= V_j ({\bf t}_j - {\bf t}_i)_{\times} V_i^T V_i^{-T} ({\bf t}_i - {\bf t}_1) \\
&= V_{j}({\bf t}_j - {\bf t}_i) \times ({\bf t}_i - {\bf t}_1) \\
&= V_j((\alpha_j - \alpha_i)({\bf t}_i - {\bf t}_1)) \times ({\bf t}_i - {\bf t}_1)) = 0.
\end{align*}
$\Leftarrow$ Without loss of generality, we assume that $i \neq 1$. Therefore,   $\exists {\bf e} \in \Real^3, {\bf e} \neq 0$ s.t $\forall j~F_{ji} {\bf e} = 0$. Since $F_{1i}=V_1[{\bf t}_1-{\bf t}_i]_{\times}V_i^T \Rightarrow V_i^{-T} ({\bf t}_i - {\bf t}_1)\in null(F_{1i})$. Assuming that  $rank(F_{1i})=2$ then  the dimension of $null(F_{1i})$ is 1, implying that ${\bf e}= \alpha V_i^{-T} ({\bf t}_i - {\bf t}_1)$, where $\alpha \neq 0 $ is a scalar. Now, $\forall j$ 
\begin{align*}
F_{ji}{\bf e}=0 & \Rightarrow V_j [{\bf t}_j - {\bf t}_i]_{\times} V_i^TV_i^{-T}({\bf t}_i - {\bf t}_1)=0 \\
&\Rightarrow V_j [{\bf t}_j - {\bf t}_i]_{\times} ({\bf t}_i - {\bf t}_1) =0 \\
& \Rightarrow ({\bf t}_j - {\bf t}_i) \times ({\bf t}_i - {\bf t}_1) =0 \\
& \Rightarrow \exists \alpha_j \in \Real~~~\text{s.t.}~~~{\bf t}_j - {\bf t}_i = \alpha_j ({\bf t}_i - {\bf t}_1) \\
& \Rightarrow {\bf t}_j = {\bf t}_i + \alpha_j ({\bf t}_i - {\bf t}_1)
\end{align*} 
concluding that the points are collinear.
\end{proof}

\begin{lemma}
    \label{lemma:gpsfmcorrection_ranks}
    Let $F=UV^T+VU^T$ where $\rank(F_i)=3$. If $\exists i\in [n]$  s.t $U_iV_i^T=0$  and w.l.o.g assuming that $\rank(U_i)\leq \rank(V_i)$ then it holds that either \begin{align*}
        \rank(U_i)=1, ~\rank(V_i)=2
    \end{align*}
    or
    \begin{align*}
        \rank(U_i)=0, ~\rank(V_i)=3
    \end{align*}
\end{lemma}
\begin{proof}
First, if $\rank(U_i)\geq 2$ then $\rank(V_i)\geq 2$ and it follows that  $U_iV_i^T\neq 0 $ which is a contradiction. Thus, \begin{align*}
    \rank(U_i)\leq 1.
\end{align*}
Therefore, we need to explore two cases: $\rank(U_i)=0$ (i.e., $U_i=0$) and $\rank(U_i)=1$.

\begin{enumerate}
\item
$U_i=0 \Rightarrow$  $F_i=U_iV^T +V_iU^T = V_iU^T$. Since $\rank(F_i)=3$ it yields $\rank(V_i)=3$.
\item
$\rank(U_i)=1 \Rightarrow$  $\rank(F_i) =\rank(U_iV^T +V_iU^T) \leq \rank(U_iV^T) + \rank(V_iU^T)\leq 1 + \rank(V_iU^T)$.
Since $\rank(F_i)=3$ it yields  $\rank(V_iU^T)\geq 2$ which means $\rank(V_i) \geq 2$. However, since $U_iV_i^T =0$ and $\rank(U_i)=1$, it holds that $\rank(V_i)=2$.
\end{enumerate}
\end{proof}
\begin{lemma} \cite{sengupta2017new} \label{lemma:gpsfm_correction_Q_3_3}
{\bf Ambiguity in the decomposition of $F$.} Let $F\in \mathbb{S}^n$ be an $n$-view fundamental matrix. Assuming $F$ can be written as $F=UV^T+VU^T$ with $U,V\in \mathbb{R}^{3n\times 3}$. Additionally, let $Q=\begin{bmatrix} Q_{11} & Q_{12} \\ Q_{21} & Q_{22}\end{bmatrix}\in \mathbb{R}^{6\times 6} $ with $Q_{11},Q_{12},Q_{21},Q_{22}\in \mathbb{R}^{3\times 3}$ that is subject to the following constraints:
\begin{enumerate}
    \item $Q_{12}Q_{11}^T+Q_{11}Q_{12}^T=0$, i.e., $Q_{12}Q_{11}^T$ is skew symmetric (6 constraints).
    \item $Q_{22}Q_{21}^T+Q_{21}Q_{22}^T=0$, i.e., $Q_{22}Q_{21}^T$ is skew symmetric (6 constraints).
    \item $Q_{12}Q_{21}^T+Q_{11}Q_{22}^T=I$ (9 constraints).
\end{enumerate}
It then follows that $F$ can be decomposed as  $F=\tilde{U}\tilde{V}^T+\tilde{V}\tilde{U}^T$ where $[\tilde{U} \  \tilde{V}]=[U \ V]Q$.
\end{lemma}

\begin{proof}
Denote by $J=\begin{bmatrix}0 &I  \\I & 0\end{bmatrix}\in \mathbb{R}^{6\times 6}$, then, 
\begin{align*}
QJQ^T=
\begin{bmatrix} Q_{11} & Q_{12} \\ Q_{21} & Q_{22}\end{bmatrix} \begin{bmatrix}0 &I  \\I & 0\end{bmatrix}\begin{bmatrix} Q_{11} & Q_{12} \\ Q_{21} & Q_{22}\end{bmatrix}^T=\end{align*}\begin{align*}
\begin{bmatrix}Q_{12} & Q_{11} \\ Q_{22} &Q_{21}\end{bmatrix}\begin{bmatrix} Q_{11}^T & Q_{21}^T \\ Q_{12}^T & Q_{22}^T\end{bmatrix}=\end{align*}\begin{align*}\begin{bmatrix}Q_{12}Q_{11}^T+Q_{11}Q_{12}^T & Q_{12}Q_{21}^T+Q_{11}Q_{22}^T \\ Q_{22}Q_{11}^T+Q_{21}Q_{12}^T & Q_{22}Q_{21}^T+Q_{21}Q_{22}^T\end{bmatrix}=\begin{bmatrix}0 &I  \\I & 0\end{bmatrix}.\end{align*}
where the rightmost equality is obtained by using the constraints on Q.
Next, let $\tilde{U},\tilde{V}\in\mathbb{R}^{3n\times 3}$ defined as $[\tilde{U} \  \tilde{V}]=[U \ V]Q$. Then, it follows that: \begin{align*}F=UV^T+VU^T=[U \ V]J[U \ V]^T=\end{align*}\begin{align*}[U \ V]QJQ^T[U \ V]^T=[\tilde{U} \ \tilde{V}]J[\tilde{U} \ \tilde{V}]^T\end{align*}\begin{align*}
=\tilde{U}\tilde{V}^T+\tilde{V}\tilde{U}^T,\end{align*}
concluding the proof.
\end{proof}
\begin{lemma}
\label{lemma:gpsfm_correction_main}
Let $F\in \mathbb{S}^n$ be an $n$-view matrix with $rank(F_i)=3$. Assuming  $F$ can be written as $F=UV^T+VU^T$ with $U,V\in \mathbb{R}^{3n\times 3}$. If $\exists i\in [n]$ such that $U_iV_i^T=0$, then there exists $Q\in \mathbb{R}^{6\times6}$ such that $[\tilde{U}, \tilde{V}] :=[U, V]Q$ where the following holds
\begin{enumerate}
    \item $F=\tilde{U}\tilde{V}^T+\tilde{V}\tilde{U}^T$.
    \item $\tilde{U}_i\tilde{V}_i^T\neq 0$.
    \item For all $j\in [n]$,  $U_jV_j^T\neq 0 \Rightarrow \tilde{U}_j\tilde{V}_j^T\neq 0$.
\end{enumerate}
\end{lemma}
\begin{proof}
Assuming w.l.o.g that $i=1$, i.e, $U_1V_1^T=0$ and $\rank(U_1) \leq \rank(V_1)$. By Lemma \ref{lemma:gpsfmcorrection_ranks}
it follows that either $\rank(U_1)=0$ and $\rank(V_1)=3$ or $\rank(U_1)=1$ and $\rank(V_1)=2$. 

We define $Q\in \mathbb{R}^{6\times 6}$ as follows
\begin{align*}
    Q_{11}=I, ~Q_{12}=0, ~Q_{22}=I, ~Q_{21}=[\vt]_\times, 
\end{align*}
with $\vt\in \Real^3$ to be defined later. 
By construction $Q$ satisfies the constraints in Lemma \ref{lemma:gpsfm_correction_Q_3_3}
and therefore for all $\vt\in \mathbb{R}^3$ if we let $[\tilde{U}, \tilde{V}] :=[U, V]Q$,
then condition 1 is satisfied. 

For all $1\leq i \leq n $
\begin{eqnarray*}
    \tilde{U}_i&=&U_i+V_i[\vt]_\times\\
    \tilde{V}_i&=&V_i,
\end{eqnarray*}
and in particular for $i=1$
\begin{align*}
    \tilde{U}_1\tilde{V}_1^T=(U_1+V_1[\vt]_\times )V_1^T=\end{align*}\begin{align*}U_1V_1^T+V_1[\vt]_\times V_1^T=V_1[\vt]_\times V_1^T.
\end{align*}
We proceed by choosing an appropriate $\vt\in \mathbb{R}^3$ such that conditions 2,3 will be satisfied. 

For the first case, i.e.,   $\rank(U_1)=0$ and $\rank(V_1)=3$, we set $\vt=\alpha (1,1,1)^T$ where $\alpha \in \mathbb{R} \backslash \{0\}$, yielding \begin{align*}
     \tilde{U}_1\tilde{V}_1^T=V_1[\vt]_\times V_1^T\neq 0
 \end{align*}
 since $\rank(V_1)=3$. This means that condition 2 is satisfied.

For the second case, i.e.,  $\rank(U_1)=1$ and  $\rank(V_1)=2$, we set $\vt=\alpha\cdot  null(V_1)$, $\alpha \in \mathbb{R} \backslash \{0\}$. Then, for the $i^{th},j^{th}$ element of $\tilde{U}_1\tilde{V}_1^T$ it holds that
\begin{align*}
(\tilde{U}_1\tilde{V}_1^T)_{ij}=(V_1[\vt]_\times V_1^T)_{ij}=\vv_i^T[\vt]_\times \vv_j=\end{align*}\begin{align*}\vv_i^T(\vt\times \vv_j)=\vt^T(\vv_j \times \vv_i),
\end{align*}
where we denote by $\vv_i^T$ the $i^{th}$ row of $V_1$.
Since  $\rank (V_1)=2$ there exist two indices $i\neq j$ where $1\leq i,j\leq 3$ s.t. $\vv_i$ and $\vv_j$ are linearly independent and span the row space of $V_1$. Using the fundamental relation in Algebra for $V_1$, i.e., $Im(V_1^T)=null(V_1)^\bot$,
it yields that $\vv_i\times \vv_j$ is in the null space of $V_1$. Therefore, $\forall \alpha \in \mathbb{R} \backslash \{0\}$ \begin{align*}  \vt^T(\vv_i\times \vv_j)= \alpha (\vv_i\times \vv_j)^T(\vv_i\times \vv_j)=\end{align*}\begin{align*} \alpha ||\vv_i\times \vv_j||^2\neq 0 \Rightarrow \tilde{U_1}\tilde{V_1}^T\neq 0 \end{align*}
Therefore, condition 2 is satisfied for the second case, as well. Now, it is left to select $\alpha$ such that condition 3 is satisfied. 

Let $j>1$ be an index such that $U_jV_j^T\neq 0$. Then since $U_jV_j^T$ is skew symmetric, it holds that 
\begin{align*}
    \rank(U_jV_j^T)=2 \Rightarrow 
    \rank(V_j),\rank(U_j)\geq 2.
\end{align*} 
Since $\rank(U_j)\geq 2$ then $U_j$ has a $2\times 2$ sub-matrix whose determinant differs from zero. Since determinant is a continuous function, there exists a small $\alpha_j$ such that the matrix $U_j+V_j[\vt]_\times$ too has a $2\times 2$ sub-matrix whose determinant differs from zero. This argument implies that
\begin{align*}
    \rank(\tilde{U}_j)=\rank(U_j+V_j[\vt]_\times)\geq \rank(U_j)\geq 2
\end{align*}
In addition, by construction it holds that  $\rank(\tilde{V}_j)=\rank(V_j)\geq 2$. Since $\tilde{U}_j,\tilde{V}_j\in \mathbb{R}^{3\times 3}$ with rank $\geq 2$, it must hold that $\tilde{U}_j\tilde{V}_j^T\neq 0$. To guarantee that for all $j\in [n]$  $U_jV_j^T\neq 0 \Rightarrow \tilde{U}_j\tilde{V}_j^T\neq 0$, we set $\alpha = \min_j {\alpha_j}$.  This concludes the proof.
\end{proof}